\documentclass[journal]{IEEEtran}
\usepackage{amsmath,amsfonts}
\usepackage{algorithmic}
\usepackage{algorithm}
\usepackage{array}
\usepackage{textcomp}
\usepackage{stfloats}
\usepackage{url}
\usepackage{verbatim}
\usepackage{graphicx}
\usepackage{cite}
\usepackage{amsmath}
\usepackage{amssymb}
\usepackage{epstopdf}
\usepackage{subfigure}
\usepackage[section]{placeins}
\usepackage{graphicx}
\usepackage{float}%提供float浮动环境
\usepackage{booktabs}%提供命令\toprule、\midrule、\bottomrule
\usepackage{picins}%个人信息
\usepackage{xcolor}
\usepackage{url}
\begin{document}

\title{ Random Fourier Bias-Compensated Filter under General Adaptive Function and Its Application for Time-Series Prediction}

\author{Yi Peng, Haiquan Zhao,~\IEEEmembership{Senior Member,~IEEE,} and Jinhui Hu 
        % <-this % stops a space
\thanks{This work was partially supported by National Natural Science Foundation of China (grant: 62171388, 61871461, 61571374). Haiquan Zhao, Yi Peng, and Jinhui hu  are with the Key Laboratory of Magnetic Suspension Technology and Maglev Vehicle, Ministry of Education, School of Electrical Engineering Southwest Jiaotong University Chengdu, China.  (e-mail: $hqzhao\_swjtu@126.com$; $pengyi1007@163.com$; $jhhu\_swjtu@126.com)$)
 
Corresponding author: Haiquan Zhao.
}}% <-this % stops a space
%\thanks{Manuscript received April 19, 2021; revised August 16, 2021.}}

% The paper headers
%\markboth{Journal of \LaTeX\ Class Files,~Vol.~14, No.~8, August~2021}%
%{Shell \MakeLowercase{\textit{et al.}}: A Sample Article Using IEEEtran.cls for IEEE Journals}

%\IEEEpubid{0000--0000/00\$00.00~\copyright~2021 IEEE}
% Remember, if you use this you must call \IEEEpubidadjcol in the second
% column for its text to clear the IEEEpubid mark.

\maketitle

\begin{abstract}
Most existing nonlinear adaptive filtering algorithms only account for output noise, neglecting the fact that input noise is also prevalent in practice. Although the recently proposed bias-compensated kernel least mean square (BCKLMS) algorithm addresses input noise in the nonlinear errors-in-variables (EIV) model, it still suffers from two major limitations. First, the use of a fixed-size dictionary restricts network growth but also prevents it from fully capturing the characteristics of the input signal. Second, as an least mean square (LMS) based algorithm, it exhibits poor robustness in the presence of non-Gaussian noise in the output signal. To overcome these issues, this paper proposes the random Fourier bias-compensated filter under general adaptive function (RFFBCGA) algorithm. Within the random Fourier feature based bias-compensated (RFFBC) framework, the proposed algorithm not only maintains a fixed network structure and effectively mitigates input noise interference through the BC term, but also achieves improved characterization of the input signal. Moreover, by leveraging the flexible form of the general adaptive (GA) function, the algorithm's robustness across various noise scenarios is further enhanced. Extensive simulations, including real-world time series prediction tasks, demonstrate the superiority of the proposed method.
\end{abstract}

\begin{IEEEkeywords}
random Fourier features (RFF), errors-in-variables (EIV), non-Gaussian noise, bias-compensated.
\end{IEEEkeywords}

\section{Introduction}
\IEEEPARstart{I}{n}  many practical engineering problems, such as time series prediction and regression analysis, when faced with complex input-output relationships that cannot be directly represented by linear models, nonlinear models naturally come into consideration \cite{KAFBOOK,NAFBOOK}. In recent years, within the field of adaptive filtering, kernel methods have emerged as one of the most prevalent approaches for addressing nonlinear problems. 

The integration of kernel methods with adaptive filters has led to kernel adaptive filters (KAFs), including algorithms like kernel least mean square (KLMS) \cite{KLMS1,KLMS2,KLMS3} and kernel recursive least squares (KRLS) algorithms \cite{KRLS}. By employing a Mercer kernel, KAFs transform the original input space into an infinite-dimensional reproducing kernel Hilbert space (RKHS). This enables the efficient computation of inner products directly through kernel function evaluations. However, KAFs generally suffer from a linearly growing network structure, which leads to continuously increasing computational and memory costs \cite{KAF_SMCA}. To mitigate this issue, a series of sparsification criteria, including the approximate linear dependence (ALD) criterion \cite{ALD}, the coherence criterion (CC) \cite{KAFBOOK}, and the quantization method \cite{quan1,quan2}, have been proposed to simplify the dictionary structure. Nevertheless, these approaches still require substantial computational resources.

To address these limitations, the random
Fourier feature kernel least mean square (RFFKLMS) algorithm \cite{RFFKLMS} was developed by introducing the least mean squares (LMS) algorithm into the random Fourier feature space (RFFs). Unlike kernel methods, random mapping projects input data into a finite-dimensional RFFs via random Fourier features (RFF) \cite{RFF1,RFF2}. The network output is obtained by projecting the input into the RFFs, followed by a linear combination with a fixed set of parameters. RFF mapping has been demonstrated to effectively approximate kernel evaluation used in KAFs \cite{rfftoklms}. Compared to KLMS, RFFKLMS maintains a fixed network dimensionality and eliminates the need for sparsification, thereby significantly reducing the computational burden.

Unfortunately, since the aforementioned algorithms are based on the mean square error (MSE) criterion, which only considers the second-order moment of the error, their performance degrades significantly in the presence of non-Gaussian background noise. To address this issue, a series of robust kernel-based algorithms and robust RFF-based methods have been developed \cite{Robust1,Robust2,Robust3,Robust4,Robust5,Robust6,Robust7,Robust8}.

Furthermore, in practical applications, adaptive filters may be subject to interference from input noise, a scenario known as the errors-in-variables (EIV) model [20]. For example, when performing time series prediction with adaptive filters, if the dataset is contaminated by noise, this situation is equivalent to having noise present in both the input and output signals. Under the EIV model, the performance of LMS-type algorithms tends to degrade accordingly. In linear adaptive filtering, two main approaches have been employed to address input noise in the EIV model: the total least squares (TLS) algorithm \cite{TLS1,TLS2,TLS3} and the bias-compensated (BC) algorithm \cite{BC1,BC2,BC3,BC4,BC5,BC6}. However, the TLS algorithm is designed based on a linear model and cannot be directly extended to nonlinear settings \cite{TLS_linear}.
Based on the BC approach, the bias-compensated for kernel LMS (BCKLMS) algorithm was recently developed \cite{BCKLMS}, which currently stands as the only available solution in the field of adaptive filtering capable of addressing nonlinear EIV model problems. Nevertheless, this algorithm suffers from two major limitations. First, although the use of a fixed-size dictionary in BCKLMS significantly reduces computational complexity, it also leads to the inability to fully capture all the characteristics of the input signal. Second, as previously mentioned, algorithms developed based on the second-order moment of the error are prone to severe performance degradation in the presence of non-Gaussian noise. To our knowledge, no existing study has yet fully resolved all the issues outlined above, which motivates this work.

In this paper, we develop a novel BC cost function and propose the random Fourier bias-compensated filter under general adaptive function (RFFBCGA) algorithm. Within the RFF-based bias-compensated (RFFBC) framework, the proposed algorithm not only maintains a fixed network structure but also  better characterizes the input signal, achieving strong performance under the nonlinear EIV model. Moreover, by leveraging the flexible form of the general adaptive (GA) function , the robustness of the proposed algorithm is further enhanced across diverse noise environments. We also provide a mean and mean-square stability analysis of the algorithm and derive two key theoretical findings that simplify its derivation process. Finally, the superiority of the proposed method is validated through extensive simulations on real-world time series prediction tasks. The main contributions of
this paper are summarized as follows:

(1) This paper proposes the RFFBCGA algorithm, which demonstrates superior performance under the nonlinear EIV model compared to other competing methods.

(2) A mean and mean-square stability analysis of the RFFBCGA algorithm is presented, yielding two key findings that simplify its theoretical derivation. Both findings are corroborated by the simulation studies.

(3) The performance of all algorithms was evaluated on two real-world datasets for time series prediction, in which the training data were contaminated with additive Gaussian noise. The simulation results confirm that the proposed algorithm achieves the superior performance under these conditions compared to other competing methods.

A brief overview of the remaining parts of this paper is organized as follows. Section II introduces the foundational concepts of the nonlinear model, KAF, RFF, and the EIV model. Section III provides the definition of the GA function and presents a detailed derivation of the proposed RFFBCGA algorithm. In Section IV, a thorough analysis of the mean behavior and mean-square stability of the RFFBCGA algorithm is provided. Section V validates the effectiveness of the RFFBCGA algorithm and the theoretical findings of this work through comprehensive computer simulations. Finally, Section VI concludes the paper.

\section{Background}
In many practical applications where nonlinear input-output relationships exist, traditional adaptive filtering algorithms struggle to model such complex relationships as 
\begin{equation}\label{1}
    d_i=\vartheta (\textbf{u}_i)+v_i
\end{equation}
where $d_i$ denotes the desired output, $i$ defines discrete time, $\textbf{u}_i\in \mathbb{R}^{M\times1}$ is the input vector, nonlinear function $\vartheta: \mathbb{R}^M \rightarrow  \mathbb{R} $, and $v_i$ is the observation noise with variance $\sigma^2_o$. To address the nonlinear estimation problem (the estimation function $\vartheta(\cdot)$ from the data sequence $\{\textbf{u}_i,d_i\}_{i=1:N}$), KAF has attracted considerable attention due to its mathematical elegance. Within the framework of the kernel method, the original data is mapped to a high-dimensional feature space through the mapping relationship $\hat \vartheta(\cdot)=\textstyle\sum_{t=1}^{i-1}c_t\kappa (\cdot , \textbf{u}_t)$, where $\kappa(\cdot,\cdot)$ denotes a Mercer kernel. The most commonly used Gaussian kernel is defined as 
\begin{equation}
    \kappa(\textbf{u}_i,\textbf{u}_t)=\exp(-\frac{\left \|\textbf{u}_i-\textbf{u}_t \right \|_2^2}{2\xi^2})
\end{equation}
where $\xi$ is the kernel width. Then, we can get 
\begin{equation}\label{ker}
    \hat{\vartheta}(\textbf{u}_i)=\textstyle\sum_{t=1}^{i-1}c_t\kappa(\textbf{u}_i,\textbf{u}_t)=\textbf{c}_{i-1}\boldsymbol\zeta ^T_i
\end{equation}
where $\textbf{c}_{i-1} = [c_1, \dots, c_{i-1}]^T$ denotes the estimated coefficient vector, and ${\boldsymbol\zeta}_i = [\kappa(\textbf{u}_i, \textbf{u}_1), \dots, \kappa(\textbf{u}_i, \textbf{u}_{i-1})]^T$ represents the kernelized input vector. Evidently, the combination of kernel functions and coefficients in (\ref{ker}) generates an expanding network structure with successive data inputs, leading to a linear growth in computational complexity over time. Although existing studies have proposed various sparsification based fixed-size dictionary construction methods, these approaches exhibit two inherent limitations: first, their dictionary training requires repeated iterations with system state variations, demonstrating suboptimal performance in dynamic environments; second, they show poor adaptability to distributed network architectures \cite{KAF_SMCA, KAF_BAD}.
\subsection{Random Fourier Features}
Compared to the sparsification approach, RFF \cite{RFF1} can provide a flexible alternative to mapping the input data $\textbf{u}_i$ into a finite and high-dimensional RFFs, i.e,
\begin{equation}
    \textbf{G}:\textbf{u}_i\in \mathbb{R}^{M\times1} \rightarrow \textbf{G}(\textbf{u}_i)\in \mathbb{R}^{D\times1}
\end{equation}
where D denotes the finite dimension ($D>M$). According to Bochner's theorem \cite{rfftoklms}, if the shift-invariant kernel (i.e., the Gaussian kernel used in this paper) is appropriately scaled, it can be guaranteed that its Fourier transform $p(\boldsymbol{w})$ constitutes a proper probability distribution. Defining $G_w(\textbf{u}_i)=e^{j(\boldsymbol {w}^T\textbf{u}_i)}$, when $\boldsymbol{w}$ is sampled from $p(\boldsymbol w)$, an unbiased estimator of the $\kappa(\textbf{u}_i,\textbf{u}_t)$ can be expressed as
\begin{equation}
\begin{aligned}
    \kappa(\textbf{u}_i,\textbf{u}_t)=\kappa(\textbf{u}_i-\textbf{u}_t)&=\int\limits_{R^M}p(\boldsymbol{w})e^{-j(\boldsymbol{w}^T(\textbf{u}_i-\textbf{ u}_t)}d\boldsymbol{w}\\
    &=E_w[G_w^H(\textbf{u}_i)G_w(\textbf{u}_t)]
\end{aligned}
\end{equation}
where $(\cdot)^H$ denotes the conjugate transpose, $j=\sqrt{-1}$ and $E_w$ is the expectation operation. $p(\boldsymbol w)$ is the zero mean Gaussian distribution and covariance matrix $\frac{1}{\xi^2}\textbf{I}_M$, where $\textbf{I}_M$ is the $M \times M$ identity matrix. When mapping the input $\textbf{u}_i$ to the D-dimensional RFFs , the commonly used mapping methods include three forms: cosine, exponential, and Gaussian. Through simulation comparisons in the literature \cite{KAF_BAD}, this study adopts the cosine mapping method, and $G(\textbf{u}_i)$ is represented as
\begin{equation}\label{eq6}
    \begin{aligned}
        G(\textbf{u}_i)&=\sqrt\frac{2}{D}\cos(\boldsymbol{w}^T\textbf{u}_i+\boldsymbol{\theta} )\\
        &=\sqrt\frac{2}{D}[\cos(\boldsymbol{w}_1^T\textbf{u}_i+\theta_1),\cdots,\cos(\boldsymbol{w}_D^T\textbf{u}_i+\theta_D)]^T
    \end{aligned}
\end{equation}
where $\boldsymbol{w}=[{\boldsymbol w_1,\boldsymbol w_2,\cdots,\boldsymbol w_D}]\in \mathbb{R}^{M\times D}$, and $\boldsymbol \theta$ is drawn uniformly from $[0,2\pi]$. Then, we have
\begin{equation} \label{G}
    \kappa(\textbf{u}_i,\textbf{u}_t)\approx G(\textbf{u}_i)^TG(\textbf{u}_t)
\end{equation}
Substituting (\ref{G}) into (\ref{ker}), yields 
\begin{equation}
    \begin{aligned}
        \hat{\vartheta}(\textbf{u}_i)&=\textstyle\sum_{t=1}^{i-1}c_t\kappa(\textbf{u}_i,\textbf{u}_t)\approx\textstyle\sum_{t=1}^{i-1}c_tG(\textbf{u}_i)^TG(\textbf{u}_t)\\
        &=\boldsymbol{\Omega}_{i-1}^TG(\textbf{u}_i)
    \end{aligned}
\end{equation}
where $\boldsymbol \Omega_{i-1}\in \mathbb{R}^{D\times1}$ is a fixed-dimensional weight vector.
\subsection{EIV model}
To construct an EIV model in which both input and output are subject to noise interference, (\ref{1}) is rewritten as
\begin{equation}
\begin{aligned}
    d_i&=\vartheta (\textbf{u}_i+\boldsymbol\eta_i)+v_i\\
    &=\vartheta(\bar{\textbf{u}}_i)+v_i
\end{aligned}   
\end{equation}
where input noise $\boldsymbol{\eta}_i \in \mathbb{R}^{M \times 1}$ follows a zero mean Gaussian distributed with variance $\sigma^2_{in}$ and assumed to be independently and identically distributed.

\section{The Important Definition and Proposed Algorithm}
In this section, several key definitions of GA function are presented, and a detailed derivation of the proposed algorithm is provided. 
\subsection{Definition of GA function}
The GA function has attracted considerable research interest due to its formal flexibility \cite{GA}. Its general form is defined as
\begin{equation}
    J_{GA}(e_i,l,\delta )=\frac{\left |\delta-2 \right |}{\delta} \left[(\frac{(e_i/l)^2}{\left |\delta-2 \right |}+1)^{\delta/2}-1\right]
\end{equation}
where $l>0$ is scale parameter, $\delta\in \mathbb{R}$ is shape parameter,  and $e_i=d_i-\hat \vartheta(\bar{\textbf{u}}_i)$.

Next, we present several special forms of the GA function.
\subsubsection{$\delta \rightarrow 2$}
\begin{equation}
    \lim\limits_{\delta\to2}\frac{\left |\delta-2 \right |}{\delta} (\frac{(e_i/l)^2}{\left |\delta-2 \right |}+1)^{\delta/2}-\frac{\left |\delta-2 \right |}{\delta}=\frac{1}{2}(e_i/l)^2
\end{equation}

\subsubsection{$\delta \rightarrow 0$}
\begin{equation}
    \lim\limits_{\delta\to0}\frac{\left |\delta-2 \right |}{\delta} (\frac{(e_i/l)^2}{\left |\delta-2 \right |}+1)^{\delta/2}-\frac{\left |\delta-2 \right |}{\delta}=\log(\frac{1}{2}(e_i/l)^2+1)
\end{equation}

\subsubsection{$\delta \rightarrow -\infty$}
\begin{equation}
    \lim\limits_{\delta\to -\infty}\frac{\left |\delta-2 \right |}{\delta} (\frac{(e_i/l)^2}{\left |\delta-2 \right |}+1)^{\delta/2}-\frac{\left |\delta-2 \right |}{\delta}=1-\exp(-\frac{1}{2}(e_i/l)^2)
\end{equation}

\subsubsection{$\delta \rightarrow +\infty$}
\begin{figure}[t]
    \centering
    \subfigure[]{
    	\begin{minipage}[b]{0.45\linewidth}%占比
        \centering
        \includegraphics[scale=0.32]{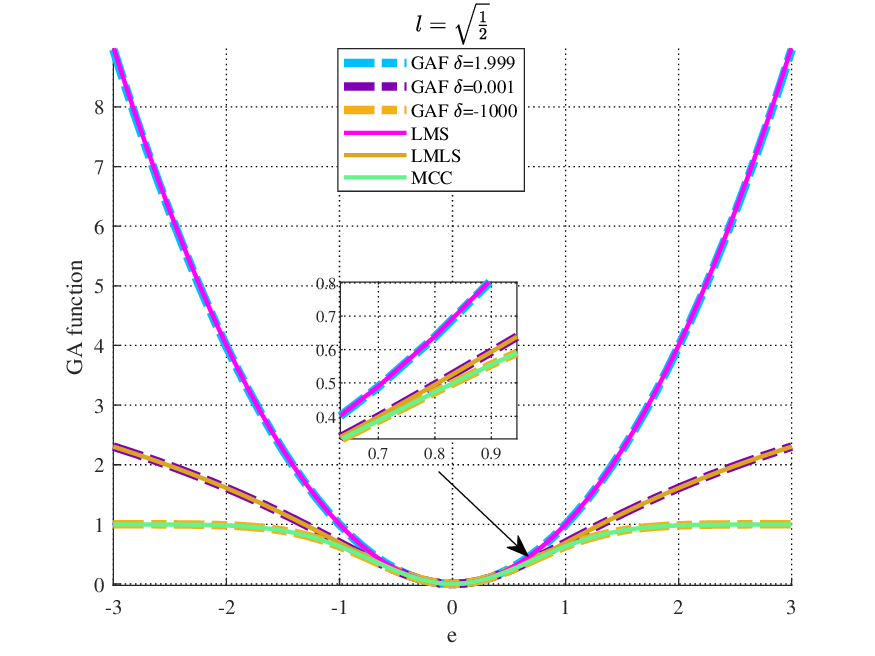} %大小 
    \end{minipage}
    }
        \subfigure[]{
    	\begin{minipage}[b]{0.48\linewidth}
        \centering
        \includegraphics[scale=0.32]{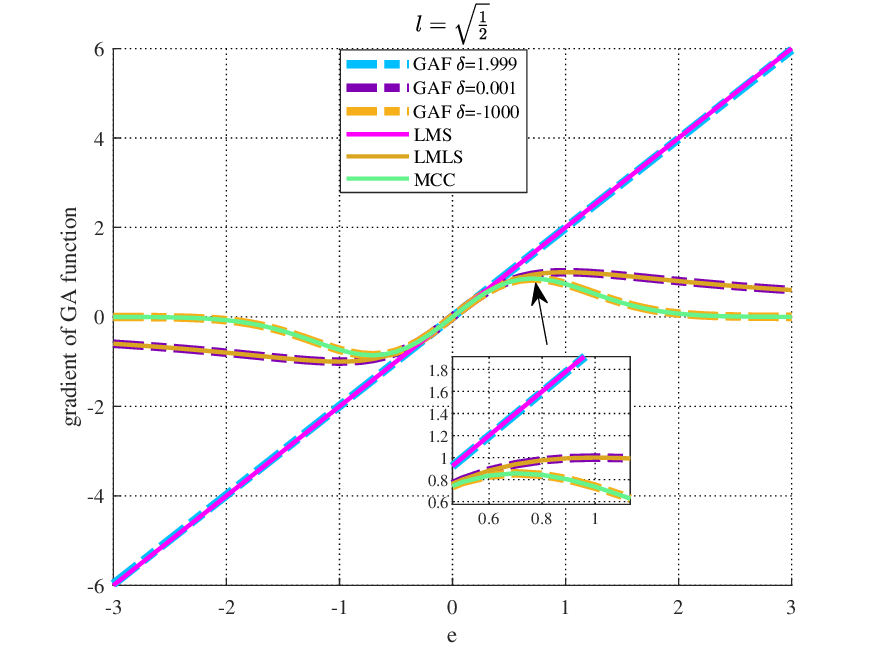}
    \end{minipage}
    }
    \caption{Curves for different cost functions and gradients}
    \label{Curves}
\end{figure}

\begin{equation}
    \lim\limits_{\delta\to -\infty}\frac{\left |\delta-2 \right |}{\delta} (\frac{(e_i/l)^2}{\left |\delta-2 \right |}+1)^{\delta/2}-\frac{\left |\delta-2 \right |}{\delta}=\exp(\frac{1}{2}(e_i/l)^2)-1
\end{equation}
$Remark $ 1: It is evident that with proper selection of the parameter $\delta$, the GA function can readily degenerate into various  common cost functions in the field of adaptive filtering: Eq. (11) for LMS \cite{LMS}, Eq. (12) for least mean logarithmic square (LMLS) \cite{LMLS}, and Eq. (13) for maximum
correntropy criterion (MCC)  \cite{MCC} algorithms. To visually demonstrate this property, Fig. \ref{Curves} shows that when appropriate parameters are selected for the GA function, it can accurately fit the LMS, LMLS, and MCC algorithms, respectively. This indicates that by adopting different parameter values, the GA function can adapt its functional shape to accommodate varying noise conditions, thereby laying the foundation for superior algorithm performance across diverse noise environments. Furthermore, since $\delta$ tends to infinity, the algorithm experiences convergence difficulties, and thus this case is not considered in this paper.

\subsection{The RFFBCGA algorithm}
Traditional TLS algorithms \cite{TLS_linear}, based on the assumption of the linear model, struggle to effectively handle input noise in nonlinear problems. To address this issue, this paper proposes a bias-compensated approach to mitigate the impact of noisy inputs on model performance. Inspired by reference \cite{BCKLMS}, this paper constructs the cost function of the RFFBCGA algorithm using the GA function, defined as  
\begin{equation}\label{J1}
\begin{aligned}\
    J_{GA}=E&\{\frac{\left |\delta-2 \right |}{\delta} [(\frac{(e_i/l)^2}{\left |\delta-2 \right |}+1)^{\delta/2}-1] -\gamma\frac{\sigma_{in}^4}{4}(\displaystyle\sum_{m=1}^{M}\frac{\partial^2 \hat\vartheta(\bar {\textbf{u}}_i)}{\partial u^2_m})^2\}\\
    =&E\{\frac{\left |\delta-2 \right |}{\delta} [(\frac{(e_i/l)^2}{\left |\delta-2 \right |}+1)^{\delta/2}-1]\}\\&-\gamma\frac{\sigma_{in}^4}{4}E\left\{{\Omega}_{i-1}^T(-G(\bar {\textbf{u}}_i))\odot\begin{bmatrix}
\left \|\boldsymbol{w_1} \right \|^2\\ 
\vdots \\ 
\left \|\boldsymbol{w_D} \right \|^2
\end{bmatrix}\right\}^2
 \end{aligned}   
\end{equation}
where $E(\cdot)$ is the expectation operation, $\odot$ is the Hadamard
product, $(\cdot)^T$ is the transpose sign, and $\gamma>0$ is the regularization parameter.

To simplify subsequent derivations, we define a diagonal matrix to replace the last term in (\ref{J1}) as
\begin{equation}
   \boldsymbol{W}_D=diag \left[ \left \|\boldsymbol{w_1} \right \|^2,\left \|\boldsymbol{w_2} \right \|^2,\cdots,\left \|\boldsymbol{w_D} \right \|^2\right]
\end{equation}

Then, its instantaneous gradient can be derived as %(%利用后面结论，RR是对角矩阵)
\begin{equation}
\begin{aligned}
    \hat {\textbf{g}}_{GA}\approx&-(e_i/l^2)(\frac{(e_i/l)^2}{|\delta-2|}+1)^{(\delta/2-1)}G(\bar{\textbf{u}}_i)\\
    &-\gamma\frac{\sigma^4_{in}}{2}\boldsymbol{W}_D G(\bar {\textbf{u}}_i)G(\bar {\textbf{u}}_i)^T\boldsymbol{W}_D\boldsymbol{\Omega}_{i-1}
\end{aligned}
\end{equation}

Using the gradient descent approach, the algorithm update formula can be obtained as
\begin{equation}\label{update}
\begin{aligned}    
    \boldsymbol{\Omega}_i&=\boldsymbol{\Omega}_{i-1}-\mu_{_{GA}}\hat {\textbf{g}}_{GA}\\
    &=\boldsymbol{\Omega}_{i-1}+\mu_{_{GA}}(e_i/l^2)(\frac{(e_i/l)^2}{|\delta-2|}+1)^{(\delta/2-1)}G(\bar{\textbf{u}}_i)\\
    &+\mu_{_{GA}}\gamma\frac{\sigma^4_{in}}{2}\boldsymbol{W}_D G(\bar {\textbf{u}}_i)G(\bar {\textbf{u}}_i)^T\boldsymbol{W}_D\boldsymbol{\Omega}_{i-1}
\end{aligned}    
\end{equation}
where $\mu_{_{GA}}$ is the step size.
Finally, Algorithm 1 summarizes the computation of the RFFBCGA algorithm.

\begin{algorithm}[]
	\caption{RFFBCGA}
    \label{alg:Sum}
    \begin{algorithmic}[1]
          \STATE Parameters $\gamma$, $\mu_{_{GA}}$, $\delta$, $l$, $\xi$
            \STATE Initialization $\boldsymbol \Omega(0)=\textbf{0}$, $\boldsymbol w_1, \cdots, \boldsymbol{w}_D $ are  drawn from $\mathcal{N}(0, \frac{1}{\xi^2}\textbf{I}_M)$ and $\boldsymbol \theta$ is drawn uniformly from $[0,2\pi]$ independently. 
    	
      	\FOR{$i=1,2,3 ...$}
            
                \STATE $\bar {\textbf{u}}_i={\textbf{u}}_i+\boldsymbol\eta_i$
        	\STATE $e_i=d_i-\hat \vartheta(\bar {\textbf{u}}_i)$
                \STATE $  \boldsymbol{W}_D=diag \left[ \left \|\boldsymbol{w_1} \right \|^2,\left \|\boldsymbol{w_2} \right \|^2,\cdots,\left \|\boldsymbol{w_D} \right \|^2\right]$
                \STATE $ G(\bar {\textbf{u}}_i)=\sqrt\frac{2}{D}[\cos(\boldsymbol{w}_1^T\bar {\textbf{u}}_i+\theta_1),\cdots,\cos(\boldsymbol{w}_D^T\bar {\textbf{u}}_i+\theta_D)]^T$
                %\STATE $Q(e_i)=(\frac{(e_i/l)^2}{|\delta-2|}+1)^{(\delta/2-1)}$
                \STATE $\begin{aligned}[t]
        \hat{\textbf{g}}_{_{GA}} \approx & -(e_i/l^2) (\frac{(e_i/l)^2}{|\delta-2|}+1)^{(\delta/2-1)} G(\bar{\mathbf{u}}_i) \\
        & -\gamma\frac{\sigma^4_{in}}{2}\boldsymbol{W}_D G(\bar {\textbf{u}}_i)G(\bar {\textbf{u}}_i)^T\boldsymbol{W}_D\boldsymbol{\Omega}_{i-1}
   \end{aligned}$
      	  \STATE $\boldsymbol{\Omega}_i=\boldsymbol{\Omega}_{i-1}-\mu_{_{GA}}\hat {\textbf{g}}_{_{GA}}$
            \ENDFOR
    
    \end{algorithmic}
\end{algorithm}

\section{Performance Analysis}
In this section, the following fundamental assumptions \cite{BC5,BC6} are proposed to facilitate a detailed performance analysis for the RFFBCGA algorithm.

A1: The weight error vector $\Delta \boldsymbol\Omega_i\triangleq\boldsymbol{\Omega}_o-\boldsymbol{\Omega}_i$ is uncorrelated with $d_i$ and $G(\bar{\textbf{u}}_i)$.

A2: The filter is long enough so that $Q(e_i)$ is uncorrelated with $G(\bar {\textbf{u}}_i)$.

A3: Both the input noise $\boldsymbol \eta_i$ and the output noise $v_i$ are zero-mean Gaussian noise. In addition, $\boldsymbol \eta_i$, $v_i$, and $\textbf{u}_i$ are mutually independent.

A4: The step-size $\mu_{_{GA}}$ is sufficiently small so that the input vector $G(\bar {\textbf{u}}_i)$ and the weight vector $\boldsymbol{\Omega}_i $ are mutually independent.

Before analyzing the mean stability of the algorithm, we first present several important conclusions and provide their proofs.

$Theorem$ 1: We give the following definition:  $\textbf{R}_i\triangleq E[G({\textbf{u}}_i)G({\textbf{u}}_i)^T]$, where ${\textbf{u}}_i$ is the noise-free input signal. Since ${\textbf{u}}_i$, $\boldsymbol w$, and $\boldsymbol \theta$ are independent of each other, and using the definition of the trigonometric function, we can get
\begin{equation}
    \textbf{R}_i=\frac{1}{D}\textbf{I}_{D\times D}
\end{equation}
$Proof$:
From (\ref{eq6}), $\textbf{R}_i$ can be derived as
\begin{equation}\label{r20}
\begin{aligned}
    \textbf{R}_i &=E[G({\textbf{u}}_i)G({\textbf{u}}_i)^T]\\
    &=E\left\{\sqrt\frac{2}{D}[\cos(\boldsymbol{w}_1^T\textbf{u}_i+\theta_1),\cdots,\cos(\boldsymbol{w}_D^T\textbf{u}_i+\theta_D)]^T \right.\\
    &\left.\times\sqrt\frac{2}{D}[\cos(\boldsymbol{w}_1^T\textbf{u}_i+\theta_1),\cdots,\cos(\boldsymbol{w}_D^T\textbf{u}_i+\theta_D)]\right\}\\
    &=\begin{bmatrix}
\textbf{r}_{11} & \cdots &\textbf{r}_{1D}  \\ 
 \vdots&\ddots  & \vdots\\ 
 \textbf{r}_{D1}&\cdots  & \textbf{r}_{DD}
\end{bmatrix}
\end{aligned}
\end{equation}
where 
\begin{equation}
    \textbf{r}_{mn}=\frac{2}{D}E\left[\cos(\boldsymbol{w}_m^T\textbf{u}_i+\theta_m)\cos(\boldsymbol{w}_n^T\textbf{u}_i+\theta_n)\right]
\end{equation}

\begin{itemize}
\item When $m = n$:
    \begin{equation}\label{item1}
    \begin{aligned}
        \textbf{r}_{mn}&=\frac{2}{D}E(\cos^2(\boldsymbol{w}_m^T\textbf{u}_i+\theta_m))\\
        &=\frac{1}{D}E(1+\cos(2\boldsymbol{w}_m^T\textbf{u}_i+2\theta_m))\\
        &=\frac{1}{D}+\frac{1}{2\pi}\int_{\boldsymbol{w}}\int_{0}^{2\pi}\cos(2\boldsymbol{w}_m^T\textbf{u}_i+2\theta_m))d\theta d\boldsymbol{w}\\
        &=\frac{1}{D}
    \end{aligned}
    \end{equation}

\item When $m\neq n$
    \begin{equation}\label{item2}
    \begin{aligned}
     \textbf{r}_{mn}&=\frac{2}{D}E(\cos(\boldsymbol{w}_m^T\textbf{u}_i+\theta_m)\cos(\boldsymbol{w}_n^T\textbf{u}_i+\theta_n))\\
     &=\frac{2}{D}E\left\{\frac{1}{2}[\cos(\boldsymbol{w}_m^T\textbf{u}_i+\boldsymbol{w}_n^T\textbf{u}_i+\theta_m+\theta_n)\right.\\
&\left.\cos(\boldsymbol{w}_m^T\textbf{u}_i-\boldsymbol{w}_n^T\textbf{u}_i+\theta_m-\theta_n)]\right\}\\
&=0
     \end{aligned}
    \end{equation}
\end{itemize}

Substituting (\ref{item1}) and (\ref{item2}) into (\ref{r20}), yields
\begin{equation}
    \textbf{R}_i=\frac{1}{D}\textbf{I}_{D\times D}
\end{equation}

$Theorem$ 2: We give the following definition:  $ \bar {\textbf{R}}_i\triangleq E[G({\bar {\textbf{u}}}_i)G({\bar {\textbf{u}}}_i)^T]$, where ${\bar {\textbf{u}}}_i$ is the noisy input signal. 
Since $\bar {\textbf{u}}_i$, $\boldsymbol w$, and $\boldsymbol \theta$ are independent of each other, and using the definition of the trigonometric function, we can get
\begin{equation}
    \bar {\textbf{R}}_i=\frac{1}{D}\textbf{I}_{D\times D}
\end{equation}
$Proof$: 
From (\ref{eq6}), $\bar {\textbf{R}}_i$ can be derived as
    \begin{equation}
    \begin{aligned} \label{cos RR}
        \bar {\textbf{R}}_i&=E[G(\bar {\textbf{u}}_i)G(\bar {\textbf{u}}_i)^T]\\
        &=E\left\{\sqrt\frac{2}{D}[\cos(\boldsymbol{w}_1^T \bar {\textbf{u}}_i+\theta_1),\cdots,\cos(\boldsymbol{w}_D^T\bar {\textbf{u}}_i+\theta_D)]^T\right.\\
        & \left.\times\sqrt\frac{2}{D}[\cos(\boldsymbol{w}_1^T\bar {\textbf{u}}_i+\theta_1),\cdots,\cos(\boldsymbol{w}_D^T\bar {\textbf{u}}_i+\theta_D)]\right\}\\
        &=E\begin{bmatrix}
\bar {\textbf{r}}_{11} & \cdots &\bar {\textbf{r}}_{1D}  \\ 
 \vdots&\ddots  & \vdots\\ 
 \bar{\textbf{r}}_{D1}&\cdots  & \bar{\textbf{r}}_{DD}
\end{bmatrix}
    \end{aligned}
\end{equation}

Subsequently, by applying the definition of the trigonometric function, $\bar {\textbf{r}}_{mn}$ can be expressed as
\begin{equation}\label{rmn1}
    \bar {\textbf{r}}_{mn}=\frac{2}{D}E\left[\textbf{P}_{1}-\textbf{P}_{2}-\textbf{P}_{3}+\textbf{P}_{4} \right]
\end{equation}
where
\begin{equation}
\begin{aligned} \label{ppp}
    &\textbf{P}_{1}=\cos (\boldsymbol{w}_m^T\textbf{u}_i+\theta_m)\cos(\boldsymbol{w}_m^T\boldsymbol{\eta}_i)\cos(\boldsymbol{w}_n^T\textbf{u}_i+\theta_n)\cos(\boldsymbol{w}_n^T\boldsymbol{\eta}_i)
\\
    &\textbf{P}_{2}=\cos (\boldsymbol{w}_m^T\textbf{u}_i+\theta_m)\cos(\boldsymbol{w}_m^T\boldsymbol{\eta}_i)\sin(\boldsymbol{w}_n^T\textbf{u}_i+\theta_n)\sin(\boldsymbol{w}_n^T\boldsymbol{\eta}_i)
\\
    &\textbf{P}_{3}=\sin (\boldsymbol{w}_m^T\textbf{u}_i+\theta_m)\sin(\boldsymbol{w}_m^T\boldsymbol{\eta}_i)\cos(\boldsymbol{w}_n^T\textbf{u}_i+\theta_n)\cos(\boldsymbol{w}_n^T\boldsymbol{\eta}_i)\\
    &\textbf{P}_{4}=\sin (\boldsymbol{w}_m^T\textbf{u}_i+\theta_m)\sin(\boldsymbol{w}_m^T\boldsymbol{\eta}_i)\sin(\boldsymbol{w}_n^T\textbf{u}_i+\theta_n)\sin(\boldsymbol{w}_n^T\boldsymbol{\eta}_i)
    \end{aligned}
\end{equation}

Under assumption A3, we take the expectation of Eq. (\ref{ppp}), leads to the following results:
\begin{itemize}
    \item  When $m$ = $n$, Eq. (\ref{ppp}) can be rewritten as
\begin{equation}
    \begin{aligned}\label{m=n1}
        &E(\textbf{P}_{1}+\textbf{P}_{4})= \frac{1}{2}\\
        &E(\textbf{P}_{2})=E(\textbf{P}_{3})=0\\
    \end{aligned}
\end{equation}
\item When $m$ $\ne$ $n$, Eq. (\ref{ppp}) can be rewritten as
\begin{equation}
\label{m!=n1}
        E(\textbf{P}_{1})=E(\textbf{P}_{2})=E(\textbf{P}_{3})=E(\textbf{P}_{4})=0
\end{equation}
\end{itemize}

Substituting (\ref{item1}), (\ref{item2}), (\ref{m=n1}) and (\ref{m!=n1}) into (\ref{cos RR}) gives
\begin{equation}
    \bar {\textbf{R}}_i=\frac{1}{D}\textbf{I}_{D\times D}
\end{equation}

$Remark$ 2: From another perspective, $Theorem $ 2 can also be easily verified. This is because, upon closer examination of Eq. (\ref{item1}) and Eq. (\ref{item2}) from the proof of $Theorem $ 1, it is evident that the terms equal to zero primarily result from the integral of the cosine function being zero, owing to the domain of the parameter $\theta$ ranging from 0 to $2\pi$. Consequently, variations in the input ${\textbf{u}}_i$ or the noisy input $\bar {\textbf{u}}_i$ will not affect the outcome of $Theorem $ 1. Moreover, we will validate $Theorems$ 1 and 2 in the simulation section.

\iffalse
$Theorem $ 3: Based on Theorem 2, we can derive the following result:
\begin{equation}
\begin{aligned}
   \bar{\mathcal{R}}_i&=E(G(\bar {\textbf{u}}_i)G(\bar {\textbf{u}}_i)^TG(\bar {\textbf{u}}_i)G(\bar {\textbf{u}}_i)^T)\\
   &=\frac{D+\frac{1}{2}}{D^2}\textbf{I}_{D\times D}
\end{aligned}
\end{equation}
$Proof$:

First, we define $\phi_m=\cos(\boldsymbol{w}_m^T \bar {\textbf{u}}_i+\theta_m)$. On the diagonal of the matrix $\bar{\mathcal{R}}_i$, its elements can be derived as
\begin{equation}
    \begin{aligned}
        E[(\phi\phi)^2]_{mm}&=E\{[\phi_m\phi_m\textstyle\sum_{n=1}^{D}\phi_n\phi_n]\}\\
        &=\frac{3}{2D^2}+(D-1)\frac{1}{D^2}\\
        &=\frac{D+\frac{1}{2}}{D^2}
    \end{aligned}
\end{equation}
and on the off-diagonals of the matrix $\bar{\mathcal{R}}_i$, it can be derived as
\begin{equation}
     E[(\phi\phi)^2]_{mn}=E[\phi_m\phi_n\textstyle\sum_{k=1}^{D}\phi_k^2]=0
\end{equation}
with 
\begin{equation}
    E(\phi_m^3)=E(\phi_m)=0
\end{equation}

\fi

\subsection{Mean Stability}

Subtracting $\boldsymbol{\Omega}_o$ from both sides of (\ref{update}) gives
\begin{equation}\label{weight e}
\begin{aligned}
\Delta \boldsymbol\Omega_i&=\Delta \boldsymbol\Omega_{i-1}-\frac{\mu_{_{GA}}}{l^2}G(\bar {\textbf{u}}_i)G(\bar {\textbf{u}}_i)^T\Delta \boldsymbol\Omega_{i-1}Q(e_i)\\
&-\frac{\mu_{_{GA}}}{l^2}G(\bar {\textbf{u}}_i)v_iQ(e_i)\\
&-\mu_{_{GA}}\gamma\frac{\sigma^4_{in}}{2}\boldsymbol{W}_D G(\bar {\textbf{u}}_i)G(\bar {\textbf{u}}_i)^T \boldsymbol{W}_D\boldsymbol\Omega_{i-1}\\
\end{aligned}
\end{equation}
where
\begin{equation}
    Q(e_i)=\left( \frac{(e_i / l)^2}{|\delta - 2|} + 1 \right)^{(\delta/2 - 1)}
\end{equation}

Under assumptions A1-A3, taking the expectation on both sides of (\ref{weight e}), and using $Theorem$ 2 yields
\begin{equation}
\begin{aligned} \label{mean 1}
\textbf{h}_i&=\textbf{h}_{i-1}-\frac{\mu_{_{GA}}}{l^2}Q^\textbf{h}\bar {\textbf{R}}_i\textbf{h}_{i-1}\\
&-\frac{\mu_{_{GA}}}{l^2}E(G(\bar {\textbf{u}}_i))E(v_iQ(e_i))\\
&-\mu_{_{GA}}\gamma\frac{\sigma^4_{in}}{2}\boldsymbol{W}^\textbf{h}\bar {\textbf{R}}_i \boldsymbol{\Omega}_{i-1}\\
&=\textbf{h}_{i-1}-\frac{\mu_{_{GA}}}{Dl^2}Q^\textbf{h}\textbf{I}_{D \times D}\textbf{h}_{i-1}\\
&-0+\mu_{_{GA}}\gamma\frac{\sigma^4_{in}}{2D}\frac{M(M+2)}{\xi^4}\textbf{I}_{D \times D} \textbf{h}_{i-1}\\
&-\mu_{_{GA}}\gamma\frac{\sigma^4_{in}}{2D}\frac{M(M+2)}{\xi^4}\textbf{I}_{D \times D}\boldsymbol{\Omega}_o
\end{aligned}
 \end{equation}
where
\begin{equation}
    Q^\textbf{h}=E(Q(e_i))
\end{equation}
\begin{equation}
    \textbf{h}_{i-1}=E(\Delta \boldsymbol\Omega_{i-1})
\end{equation}
\begin{equation}
    \boldsymbol{W}^\textbf{h}=E(\boldsymbol{W}_D^2)=\frac{M(M+2)}{\xi^4}\textbf{I}_{D \times D}
\end{equation}

From (\ref{mean 1}), it can be observed that if the matrix
\begin{equation}
    \boldsymbol{\mathcal{H}}=\textbf{I}-\mu_{_{GA}}(\frac{ Q^\textbf{h}}{Dl^2}-\frac{\gamma\sigma_{in}^4M(M+2)}{2D\xi^4})\textbf{I}
\end{equation}
is stable, the RFFBCGA algorithm is able to maintain mean stability. According to matrix theory, this is equivalent to
\begin{equation}\label{u1}
    \left |1-\mu_{_{GA}}(\frac{2 Q^\textbf{h}\xi^4-\gamma\sigma^4_{in}l^2M(M+2)}{2D\xi^4l^2}) \right |<1
\end{equation}

Finally, the condition for mean stability of the algorithm can be derived as
\begin{equation}\label{m43}
    0<\mu_{_{GA}}<\frac{4D\xi^4l^2}{2 Q^\textbf{h}\xi^4-\gamma\sigma^4_{in}l^2M(M+2)}
\end{equation}
$Remark$ 3: From Eq. (\ref{m43}), we can see that to ensure the denominator of the formula remains positive, this algorithm cannot handle cases where the input vector dimensionality is too high or the input noise variance is too large. However, it is evident that this problem can also be mitigated by manually tuning parameters $\xi$ and $l$.

\subsection{ Mean-Square Stability}
Under assumptions A1-A3 and Theorems 1-3, multiplying Eq. (\ref{weight e}) by its transpose and taking the expectation gives
\begin{equation}
\begin{aligned} \label{mean square}
     \mathcal{F}_i&= \mathcal{F}_{i-1}-\frac{\mu_{_{GA}}}{Dl^2}Q^\textbf{h}\mathcal{F}_{i-1}
     -\frac{\mu_{_{GA}}\sigma^4_{in}\gamma M(M+2)}{2D\xi^4}\mathcal{K}_{i-1}\\
     &-\frac{\mu_{_{GA}}}{Dl^2}Q^\textbf{h}\mathcal{F}_{i-1}+\frac{\mu_{_{GA}}^2(D+\frac{1}{2})}{D^2l^4}E(Q^2(e_i))\mathcal{F}_{i-1}\\
     &+\frac{\mu_{_{GA}}^2\gamma\sigma^4_{in}M(M+2)(D+\frac{1}{2})}{2\xi^4D^2l^2}Q^\textbf{h}\mathcal{K}_{i-1}\\
     &+\frac{\mu_{_{GA}}^2}{Dl^4}E(v_i^2Q^2(e_i))\textbf{I}-\frac{\mu_{_{GA}}\sigma^4_{in}\gamma M(M+2)}{2D\xi^4}\mathcal{K}^T_{i-1}\\
     &+\frac{\mu_{_{GA}}^2\sigma^4_{in}\gamma M(M+2)(D+\frac{1}{2})}{2D^2l^2\xi^4}Q^\textbf{h}\mathcal{K}^T_{i-1}\\
     &+\frac{\mu^2_{_{GA}}\gamma^2\sigma^8_{in}M(M+2)(M+4)(M+6)(D+\frac{1}{2})}{4D^2\xi^8}\mathcal{J}_{i-1}
\end{aligned}
\end{equation}
where
\begin{equation}
    \mathcal{F}_i=E(\Delta \boldsymbol\Omega_{i}\Delta \boldsymbol\Omega_{i}^T)
\end{equation}
\begin{equation}\label{38}
\mathcal{K}_{i-1}=E(\Delta \boldsymbol\Omega_{i-1}\boldsymbol\Omega_{i-1}^T)
\end{equation}
\begin{equation}\label{39_r}
\mathcal{J}_{i-1}=E( \boldsymbol\Omega_{i-1}\boldsymbol\Omega_{i-1}^T)
\end{equation}
\begin{equation}
    E(\boldsymbol{W}_D^4)=\frac{M(M+2)(M+4)(M+6)}{\xi^8}\textbf{I}_{D \times D}
\end{equation}

%Here, since the variance of the input noise is small, we neglect the terms involving $\sigma_{in}^8$.
Using A1, Eq. (\ref{38}) and Eq. (\ref{39_r}) can be derived as
\begin{equation}
\begin{aligned}\label{39}
    \mathcal{K}_{i-1}&=E\left[\Delta \boldsymbol\Omega_{i-1}(\boldsymbol\Omega_{o}-\Delta \boldsymbol\Omega_{i-1})^T\right]\\
    &=\textbf{h}_{i-1}\boldsymbol\Omega_{o}^T-\mathcal{F}_{i-1}
    \end{aligned}
\end{equation}

\begin{equation}
\begin{aligned}\label{39_1}
    \mathcal{J}_{i-1}&=E\left[(\boldsymbol\Omega_{o}-\Delta \boldsymbol\Omega_{i-1})(\boldsymbol\Omega_{o}-\Delta \boldsymbol\Omega_{i-1})^T\right]\\
    &=\boldsymbol\Omega_{o}\boldsymbol\Omega_{o}^T-\boldsymbol\Omega_{o}\textbf{h}_{i-1}^T-\textbf{h}_{i-1}\boldsymbol\Omega_{o}^T+\mathcal{F}_{i-1}
    \end{aligned}
\end{equation}

Substituting (\ref{39}) and (\ref{39_1}) into (\ref{mean square}) obtains
\begin{equation}
\begin{aligned} \label{40}
     \mathcal{F}_i&= \mathcal{F}_{i-1}-\frac{2\mu_{_{GA}}}{Dl^2}Q^\textbf{h}\mathcal{F}_{i-1}
     \\
     &+\frac{\mu_{_{GA}}^2(D+\frac{1}{2})}{D^2l^4}E(Q^2(e_i))\mathcal{F}_{i-1}+\frac{\mu_{_{GA}}^2}{Dl^4}E(v_i^2Q^2(e_i))\textbf{I}\\
     &+\frac{\mu_{_{GA}}\sigma^4_{in}\gamma M(M+2)}{D\xi^4}\mathcal{F}_{i-1}\\
     &-\frac{\mu_{_{GA}}^2\gamma\sigma^4_{in}M(M+2)(D+\frac{1}{2})}{\xi^4D^2l^2}Q^\textbf{h}\mathcal{F}_{i-1}\\
     &+\frac{\mu^2_{_{GA}}\gamma^2\sigma^8_{in}M(M+2)(M+4)(M+6)(D+\frac{1}{2})}{4D^2\xi^8}\mathcal{F}_{i-1}\\
     &-\frac{\mu_{_{GA}}\sigma^4_{in}\gamma M(M+2)}{2D\xi^4}(\textbf{h}_{i-1}\boldsymbol\Omega_{o}^T+\boldsymbol\Omega_{o}\textbf{h}_{i-1}^T)\\
     &+\frac{\mu_{_{GA}}^2\gamma\sigma^4_{in}M(M+2)(D+\frac{1}{2})}{2\xi^4D^2l^2}Q^\textbf{h}(\textbf{h}_{i-1}\boldsymbol\Omega_{o}^T+\boldsymbol\Omega_{o}\textbf{h}_{i-1}^T)\\
     &+\frac{\mu^2_{_{GA}}\gamma^2\sigma^8_{in}M(M+2)(M+4)(M+6)(D+\frac{1}{2})}{4D^2\xi^8}\boldsymbol\Omega_{o}\boldsymbol\Omega_{o}^T\\
     &-\frac{\mu^2_{_{GA}}\gamma^2\sigma^8_{in}M(M+2)(M+4)(M+6)(D+\frac{1}{2})}{4D^2\xi^8}\\
     &\times (\textbf{h}_{i-1}\boldsymbol\Omega_{o}^T+\boldsymbol\Omega_{o}\textbf{h}_{i-1}^T)
\end{aligned}
\end{equation}

From (\ref{40}), to ensure the mean-square stability of the algorithm, the following matrix should be stable:
\begin{equation}
\begin{aligned}\label{52}
    \boldsymbol{\mathcal{P}}&= \mathcal{F}_{i-1}-\frac{2\mu_{_{GA}}}{Dl^2}Q^\textbf{h}\mathcal{F}_{i-1}
     \\
     &+\frac{\mu_{_{GA}}^2(D+\frac{1}{2})}{D^2l^4}E(Q^2(e_i))\mathcal{F}_{i-1}\\
     &+\frac{\mu_{_{GA}}\sigma^4_{in}\gamma M(M+2)}{D\xi^4}\mathcal{F}_{i-1}\\
     &-\frac{\mu_{_{GA}}^2\gamma\sigma^4_{in}M(M+2)(D+\frac{1}{2})}{\xi^4D^2l^2}Q^\textbf{h}\mathcal{F}_{i-1}\\
     &+\frac{\mu^2_{_{GA}}\gamma^2\sigma^8_{in}M(M+2)(M+4)(M+6)(D+\frac{1}{2})}{4D^2\xi^8}\mathcal{F}_{i-1}\\
    \end{aligned}
\end{equation}

Since $\mu_{_{GA}}$ is significantly less than 1, $\mu_{_{GA}}^2$ becomes negligible. Thus, we omit the terms involving $\mu_{_{GA}}^2$ in Eq. (\ref{52}), which simplifies the expression to
\begin{equation}
\begin{aligned}
            \boldsymbol{\mathcal{P}}\approx&\mathcal{F}_{i-1}-\frac{2\mu_{_{GA}}}{Dl^2}Q^\textbf{h}\mathcal{F}_{i-1}
     +\frac{\mu_{_{GA}}\sigma^4_{in}\gamma M(M+2)}{D\xi^4}\mathcal{F}_{i-1}\\
\end{aligned}
\end{equation}

Then, based on matrix theory and the properties of the Kronecker product, the condition for ensuring the stability of the matrix $\boldsymbol{\mathcal{P}}$ is derived as
\begin{equation}
    \left |1-\frac{2\mu_{_{GA}}}{Dl^2}Q^\textbf{h}+\frac{\mu_{_{GA}}\gamma\sigma^4_{in}M(M+2)}{D\xi^4} \right |<1
\end{equation}
which is equivalent to
\begin{equation}\label{47}
    0<\mu_{_{GA}}<\frac{2D\xi^4l^2}{2 Q^\textbf{h}\xi^4-\gamma\sigma^4_{in}l^2M(M+2)}
\end{equation}

$Remark$ 4: A comparison of (\ref{m43}) and (\ref{47}) shows that the maximum step-size in (\ref{47}) is half that of (\ref{m43}). Consequently, the bound in (\ref{47}) is tighter, thus ensuring convergence in both the mean and mean-square senses.

\section{Simulation}
In this section, the superiority of the proposed algorithm will be demonstrated through extensive simulations and evaluated within the context of time series prediction over 30 independent Monte
Carlo (MC) runs. Additionally, the derived $Theorems$ 1 and 2 will be verified through simulations. Given the scarcity of existing nonlinear adaptive filtering algorithms that can effectively handle input noise, our simulation study will focus on a comparative analysis with the BCKLMS algorithm. The nonlinear system investigated in this study is detailed in \cite{Robust5}, and we will adopt this same system in the subsequent Examples I and II. Since the traditional kernel method and the RFF method yield weight vectors of different dimensionalities, the test mean square error (MSE) is uniformly adopted as the performance indicator in this study, as defined by the following equation:
\begin{equation}
    MSE(dB)=\frac{1}{N}10\log10\textstyle\sum_{i=1}^{N}(d_i-\hat\vartheta_i)^2
\end{equation}
A total of 50100 samples are utilized, with the first 50000 noise-contaminated samples dedicated to training the adaptive filter \cite{BCKLMS}. The remaining 100 samples are then used to compute the test MSE with the trained filter. 

In addition, to evaluate the algorithm's performance under non-Gaussian noise, two common noise models are briefly described. 

(1) In the first case, the noise is modeled by a Bernoulli-Gaussian (BG) process defined as $\rho_i = \beta_i\psi _i$. In this model, $\psi_i$ is a Bernoulli process described by the probabilities $P(\psi_i=1) = p_{b}$ and $P(\psi_i=0) = 1 - p_{b}$, with $p_{b}$ representing the probability of occurrence . The component $\beta_i$ is a zero-mean white Gaussian process with variance $\sigma^2_\beta$.

(2) The second case employs impulsive noise modeled by an $\alpha$-stable process, generated from the characteristic function:
\begin{equation}
\mathcal{A}_i=\exp\left\{j(\mathcal{E}t)-z|t|^\tau\left[1+j(\mathcal{B}\text{sgn}(t)\mathcal{S}(t,\tau))\right]\right\}
\end{equation}
where $\tau \in (0, 2]$ is the characteristic factor that governs the heaviness of the tails in the distribution; $\mathcal{B} \in (-1, 1)$ measures asymmetry; $z > 0$ is the dispersion parameter; and $\mathcal{E} \in (-\infty,\infty)$ denotes the location parameter. The function $\mathcal{S}(t,\tau)$ is defined as:
\begin{equation}
    \mathcal{S}(t,\tau)=\left\{\begin{matrix}
 \tan\frac{\tau\pi}{2},& \tau \ne 1 \\ 
 \frac{2}{\pi}\log|t|& \tau =1
\end{matrix}\right.
\end{equation}
A smaller $\tau$ results in a heavier-tailed distribution, and the $\alpha$-stable distribution reduces to Gaussian distribution when $\tau = 2$. In contrast, a smaller $z$ yields fewer large outliers. The parameters of the $\alpha$-stable noise are summarized by the vector $ L = (\tau, \mathcal{B}, z , \mathcal{E})$.

\subsection{Example I (M=1)}
In this subsection, to facilitate a more effective performance comparison between the proposed algorithm and the BCKLMS algorithm, the simulation environment is configured according to reference \cite{BCKLMS}, with the specific parameters detailed below: The unknown weight vector is defined as $\textbf{w}_\Omega = [-1.5259, 0.8412, 0.2231, -0.45195, -1.2485]^T$. The dictionary $D_{c}$, with a size of $M_d = 5$, is given by $[0.7673, 0.2039, 1.2463, -0.7148, -0.2466]$. The noise-free input signal $\textbf{u}_i$ followed the standard white Gaussian distributed process. The input noise is the additive white Gaussian noise (AWGN) with signal-to-noise ratio (SNR) of 10 dB. The specific configuration of the output noise is provided in the corresponding figure caption.
\begin{figure}[t]
    \centering 
    \subfigure[]{
    	\begin{minipage}[b]{1\linewidth}%占比
        \centering
        \includegraphics[scale=0.40]{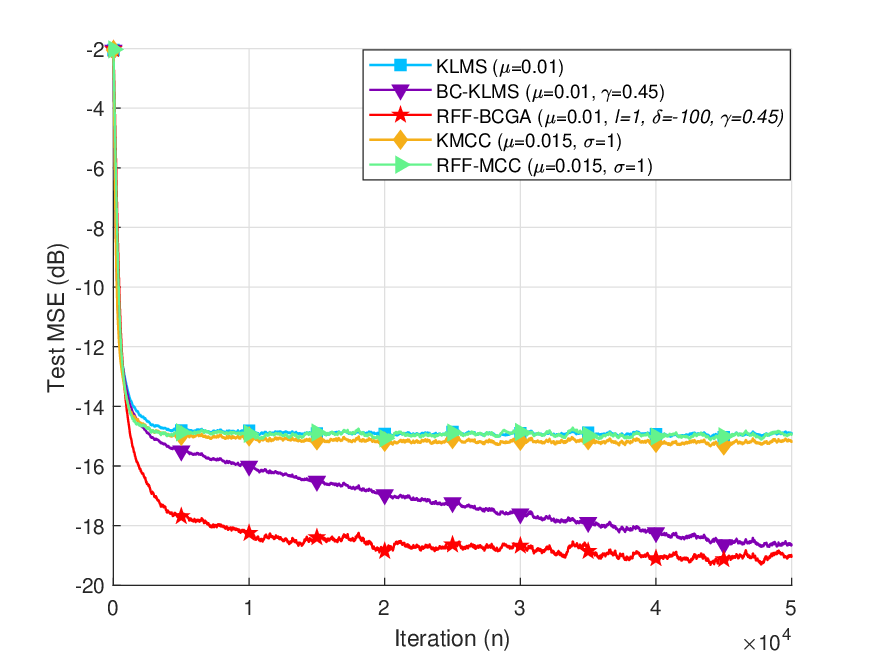} %大小
    \end{minipage}
    }
        \subfigure[]{
    	\begin{minipage}[b]{1\linewidth}
        \centering
        \includegraphics[scale=0.40]{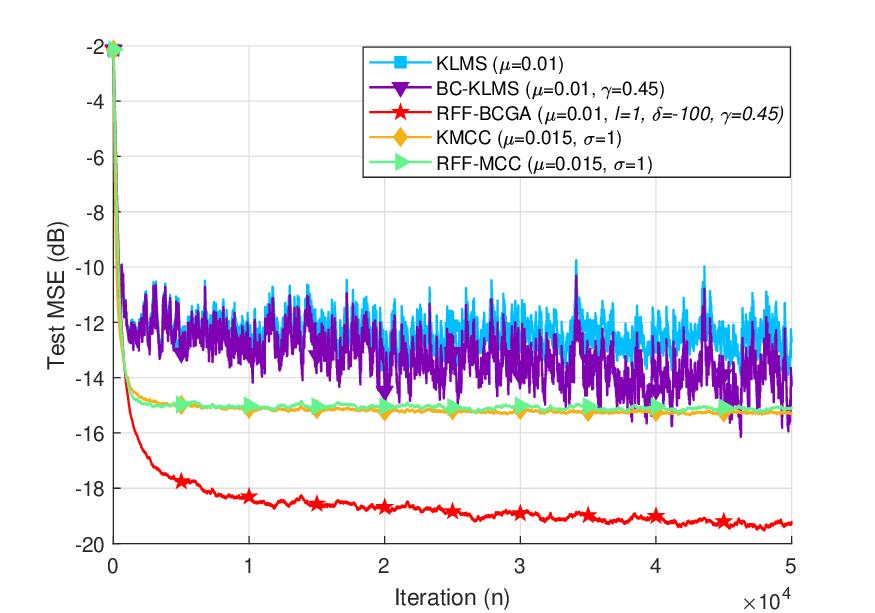}
    \end{minipage}
    }
            \subfigure[]{
    	\begin{minipage}[b]{1\linewidth}
        \centering
        \includegraphics[scale=0.40]{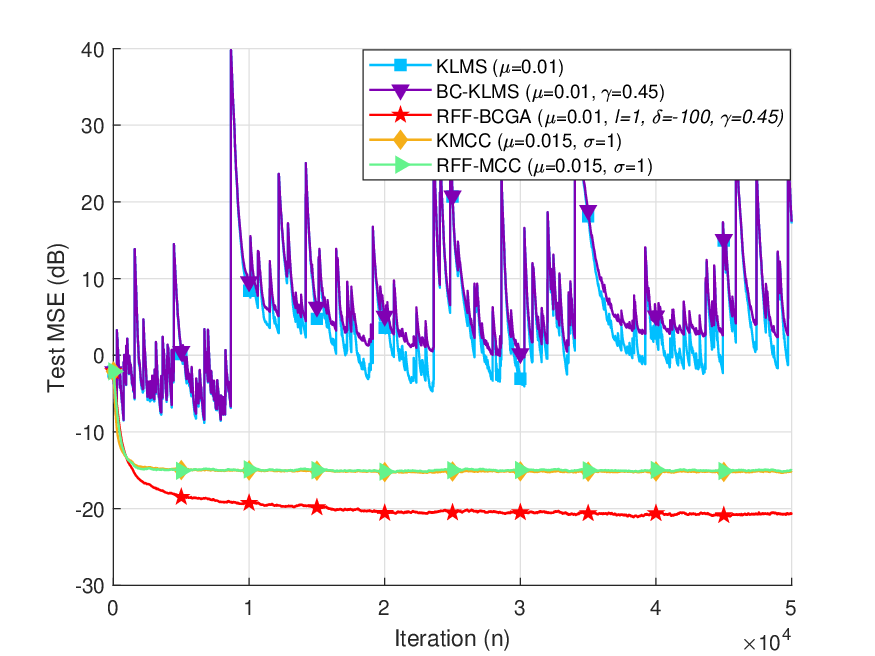}
    \end{minipage}
    }
    \caption{ Example 1. Testing MSE with the kernel bandwidth $\xi$=0.6 under different noise environment (a) AWGN with $SNR_o$=30dB; (b) 30 dB AWGN with BG noise ($p_b=0.01$, $\beta_i=500$); (c) 30dB AWGN with $\alpha$-stable noise $L=(1.0,0,0.1,0)$ }
    \label{EXP1}
\end{figure}

Fig. \ref{EXP1} illustrates the performance comparison between the proposed algorithm and various competing algorithms under different environmental noises. As shown in Fig. \ref{EXP1}(a), the KLMS, random Fourier filters under maximum
correntropy criterion (RFFMCC), and kernel adaptive filtering with maximum correntropy criterion (KMCC) algorithms maintain a high steady-state MSE because they cannot handle input noise effectively. Compared to the BCKLMS algorithm, although both eventually converge to similar steady-state MSE levels, the proposed RFFBCGA algorithm demonstrates a significantly faster convergence rate. This is attributed to the fact that while BCKLMS restricts the linear growth of computational complexity through fixed-dimensional kernel mapping, the inherent limitations of its fixed dictionary prevent it from comprehensively capturing all the characteristics of the input signal. As evidenced in Figs. \ref{EXP1}(b) and (c), when the system is disturbed by non-Gaussian noise, the BCKLMS algorithm suffers severe performance degradation due to its insufficient robustness against outliers. In contrast, the proposed RFFBCGA algorithm maintains excellent performance.

\subsection{Example II (M=2)}
The unknown weight vector of the nonlinear
system is set to $\textbf{w}_\Omega = [0.15, 0.3, 0.2, -0.15, -0.3]^T$ and the dictionary $D_c$, with a size of $M_d = 5$, is given by
\begin{equation}
D_c=\left\{\begin{bmatrix}
 0.72\\1.44 

\end{bmatrix}\begin{bmatrix}
 3.31\\1.28

\end{bmatrix}\begin{bmatrix}
 -3.03\\-2.75 

\end{bmatrix}\begin{bmatrix}
1.48\\-1.66

\end{bmatrix}\begin{bmatrix}
 -1.28\\-0.32 
\end{bmatrix}\right\}
\end{equation}
The input noise is AWGN with $SNR_i = 10$. The noise-free input
$\textbf{u}_i = [u_{1,i}, u_{2,i}]^T$ is modeled as a sequence of independent random vectors, where its components are correlated via $u_{1,i} = 0.5u_{2,i} + v_{u,i}$. Here, $u_{2,i}$ is drawn from a standard white Gaussian distribution, and $v_{u,i}$ represents zero-mean white Gaussian noise with a variance of 0.75.
\begin{figure}[htbp]
    \centering 
    \subfigure[]{
    	\begin{minipage}[b]{1\linewidth}%占比
        \centering
        \includegraphics[scale=0.40]{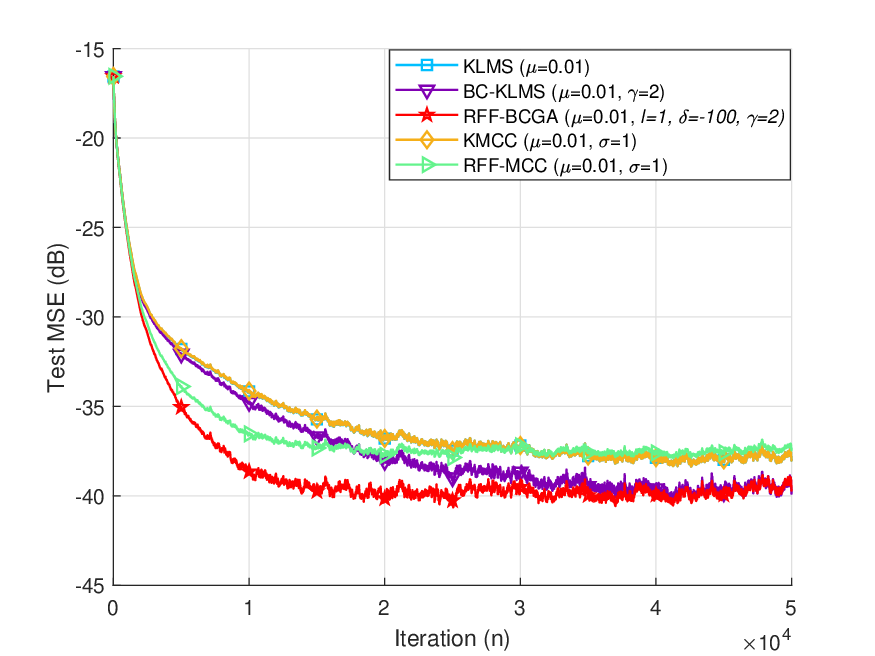} %大小
    \end{minipage}
    }
        \subfigure[]{
    	\begin{minipage}[b]{1\linewidth}
        \centering
        \includegraphics[scale=0.40]{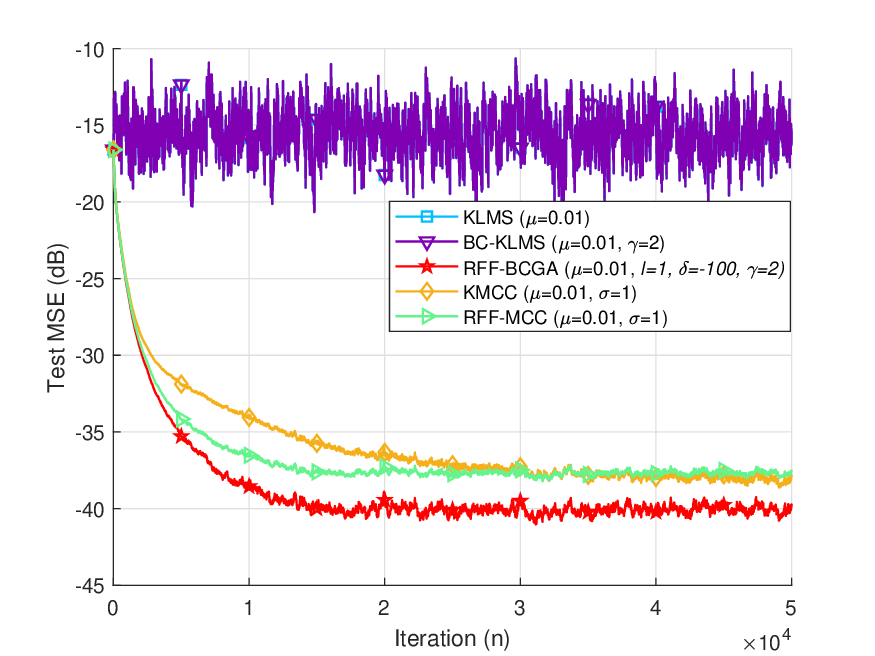}
    \end{minipage}
    }
            \subfigure[]{
    	\begin{minipage}[b]{1\linewidth}
        \centering
        \includegraphics[scale=0.40]{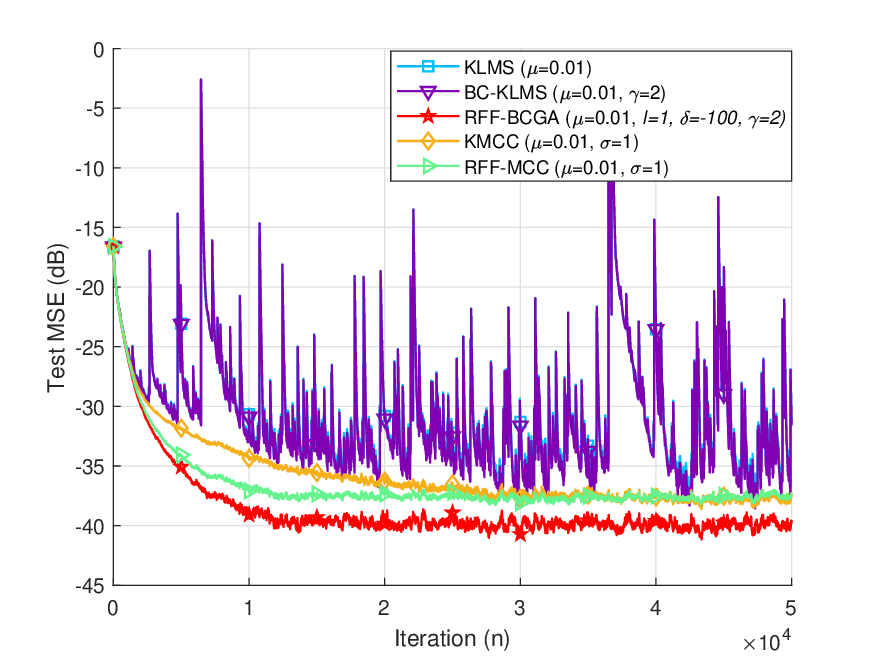}
    \end{minipage}
    }
    \caption{ Example 2. Testing MSE with the kernel bandwidth $\xi$=1.8 under different noise environment (a) AWGN with $SNR_o$ = 30dB; (b) 30 dB AWGN with BG noise ($p_b=0.01$, $\beta_i=500$); (c) 30dB AWGN with $\alpha$-stable noise $L=(1.2,0,0.01,0)$ }
    \label{EXP2}
\end{figure}
Fig. \ref{EXP2} illustrates the performance comparison between the proposed algorithm and various competing algorithms under different environmental noises. As shown in Fig. \ref{EXP2}, RFF-based algorithms demonstrate significantly faster convergence than kernel methods, which is consistent with the results in Fig. \ref{EXP1}. Furthermore, the proposed RFFBCGA algorithm maintains superior performance across various conditions.

\subsection{Time Series Prediction}
In this subsection, the first 3,000 noise-contaminated samples are used for training, with the subsequent 100 samples reserved for test MSE computation. The specific noise configuration is detailed in the corresponding figure caption. The dictionary size and input vector length are set to $M_d = 10$ and $M = 1$, respectively. It is noteworthy that, in time series prediction, since noise is directly injected into the training set, the training process must account for a scenario in which both input and output signals are contaminated. Furthermore, as the dictionary selection method is not specified in \cite{BCKLMS}, we accordingly construct the dictionary by randomly selecting $M_d$ samples from the testing set. It is also worth noting that, for clarity in the result curves, a smoothing filter with a window length of 50 is applied to the output in this section.
\subsubsection{ Sunspot Time Series Prediction}

Sunspots are visibly dark regions on the Sun's photosphere. Predicting their activity is crucial for weather forecasting, understanding conditions in Earth's atmosphere, and assessing the spatial conditions of wireless communication and broadcasting. Given its relevance, the sunspot time series has been widely used in nonlinear system modeling \cite{sunspot}. In this study, we further evaluate the proposed algorithm using the real-world monthly sunspot series (from January 1749 to January 2025), obtained from the Solar Influences Data Analysis Center\footnote{[Online Dataset]. Available: 
\url{https://www.sidc.be/silso}}.

\begin{figure}[t]
    \centering
    \subfigure[]{
    	\begin{minipage}[b]{1\linewidth}%占比
        \centering
        \includegraphics[scale=0.27]{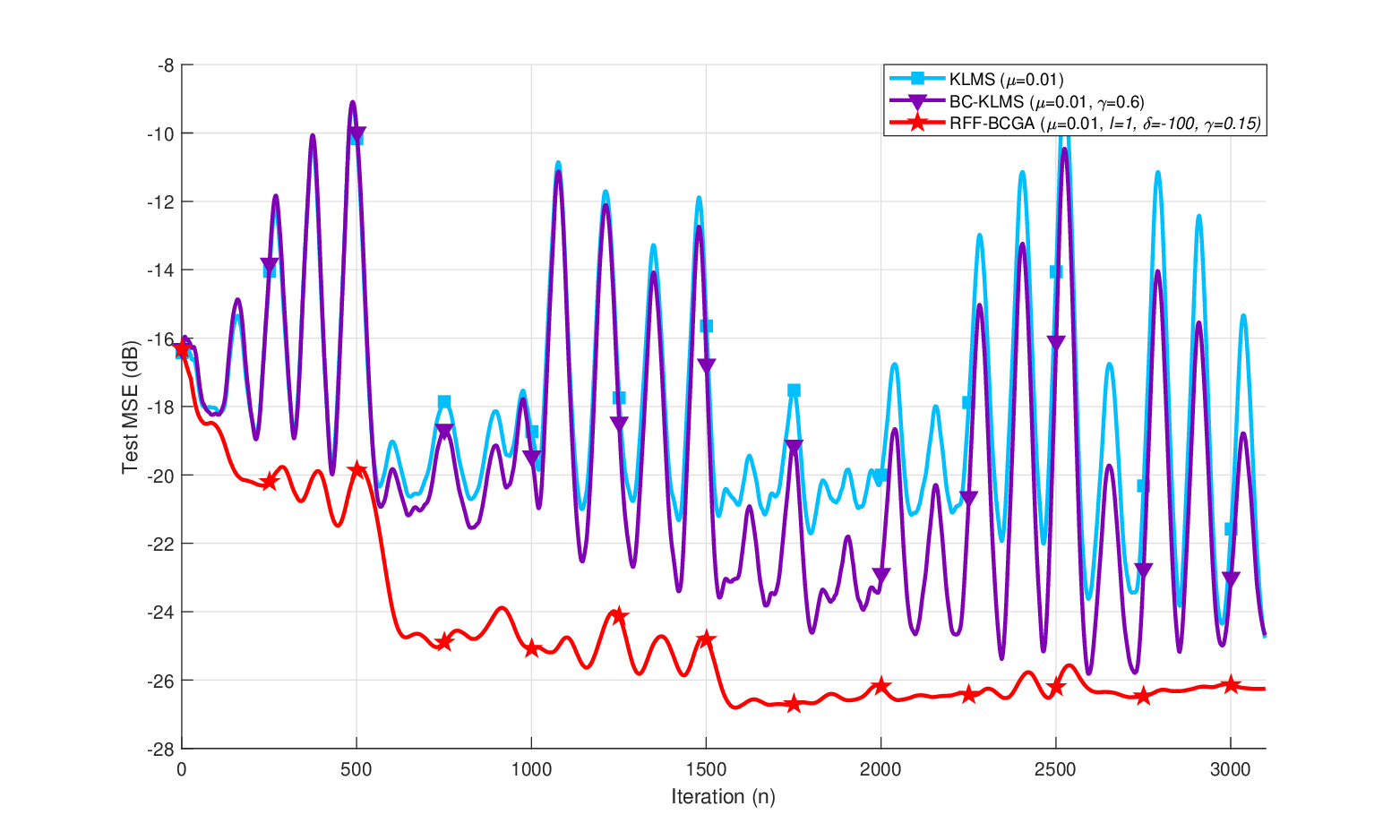} %大小 
    \end{minipage}
    }
        \subfigure[]{
    	\begin{minipage}[b]{1\linewidth}
        \centering
        \includegraphics[scale=0.27]{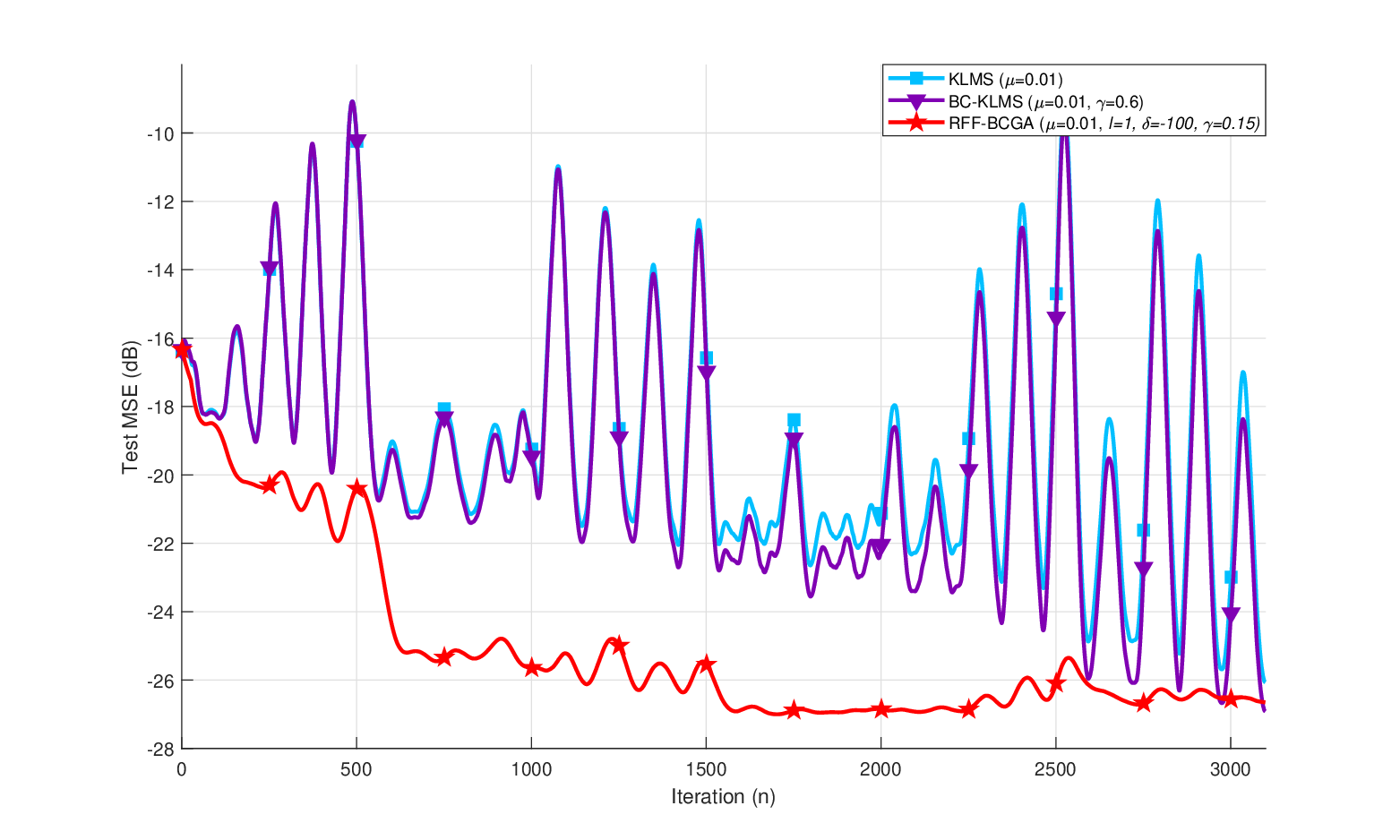}
    \end{minipage}
    }
    \caption{Testing MSE for prediction of noisy sunspots time series (kernel bandwidth $\xi=0.35$) under different noise conditions: (a) 5dB AWGN; (b) 10dB AWGN.}
    \label{SUN}
\end{figure} 

Fig. \ref{SUN} illustrates the prediction performance of the proposed and competing algorithms for the sunspot time series when the training set is contaminated by noise. As shown, both the KLMS and BCKLMS algorithms exhibit significant performance fluctuations. This is attributed to the random selection of dictionary elements from the test set, rather than employing a carefully designed dictionary. These results indicate that the performance of conventional fixed-dimensional algorithms is highly dependent on the dictionary selection, a limitation that does not affect the proposed RFFBCGA algorithm. Furthermore, the proposed algorithm maintains superior performance across varying noise levels.
\begin{figure}[htbp]
\centering
\includegraphics[scale=0.3]{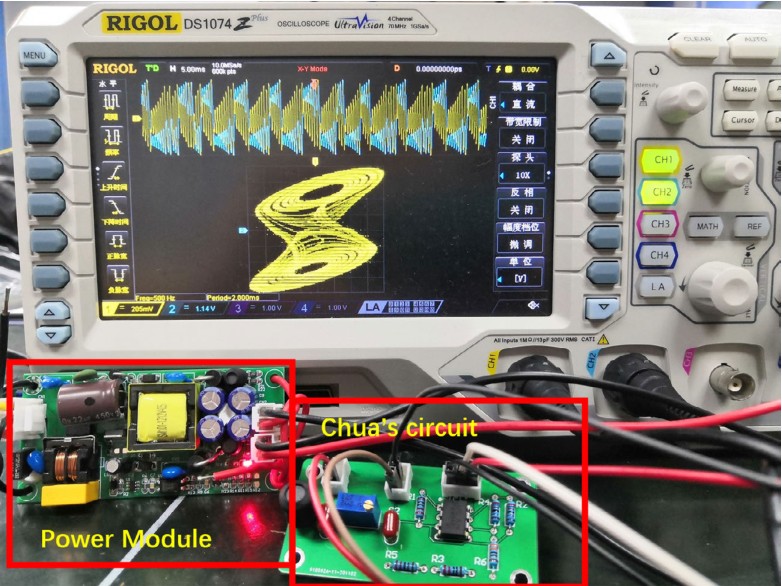}
\caption{ Chua’s circuit system}
\label{CH1}
\end{figure}\begin{figure}[htbp]
\centering
\includegraphics[scale=0.3]{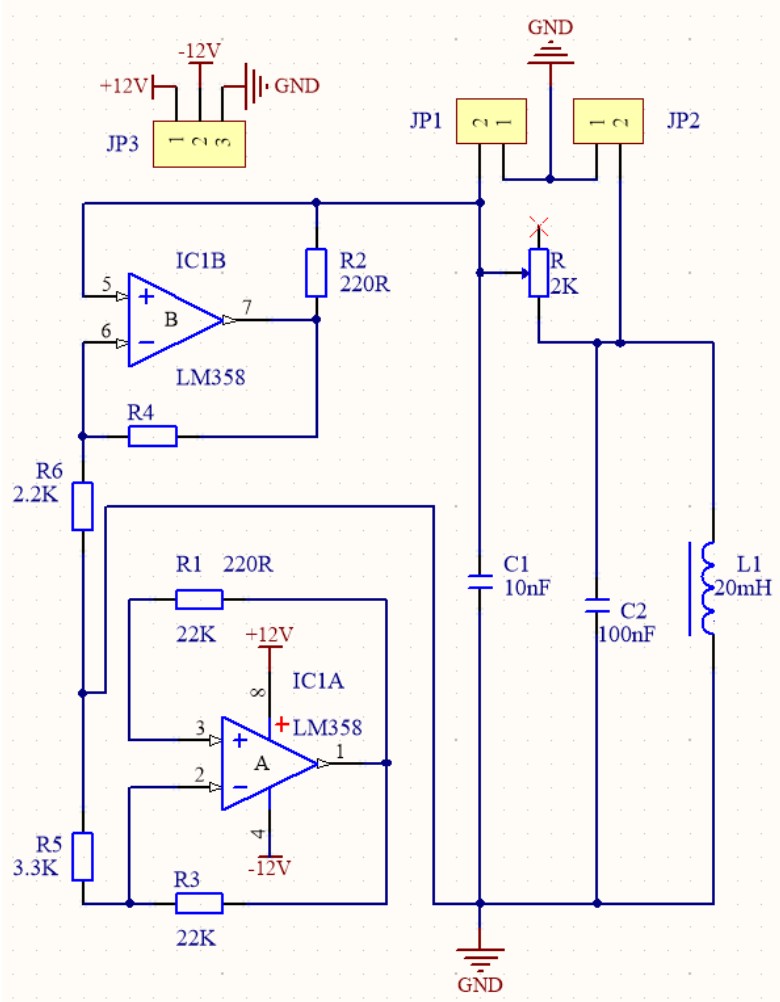}
\caption{Chua’s circuit system schematic diagram}
\label{CH2}
\end{figure}

\subsubsection{Chua’s circuit Time Series Prediction}
This subsection utilizes a physical Chua's circuit to generate chaotic time series, thereby facilitating the evaluation of the RFFBCGA algorithm's nonlinear learning capability. The circuit schematic and system implementation are shown in Figs. \ref{CH1} and \ref{CH2}, respectively. The governing equations of the Chua's circuit are given by:
\begin{equation}
\left\{
\begin{array}{l}
\dot{u}_1 = \dfrac{u_2}{RC_1} - \dfrac{u_1}{RC_1} - \dfrac{\varphi(u_1)}{C_1}, \\[10pt]
\dot{u}_2 = \dfrac{u_1}{RC_2} - \dfrac{u_2}{RC_2} + \dfrac{I_{\tilde{L}}}{C_2}, \\[10pt]
I_{\tilde{L}} = -\dfrac{u_2}{\tilde{L}},
\end{array}
\right.
\end{equation}
where \( I_{\tilde{L}} \), \( u_1 \), and \( u_2 \) represent the current through the inductor \( \tilde{L} \), and the voltages across capacitors \( C_1 \) and \( C_2 \), respectively. The function \( \varphi(u_1) \)is a segmented continuous function that describes the
diode voltage-ampere characteristic. The voltage across the capacitor \( C_1 \) is sampled to form the time series for prediction, with the latest voltage value predicted based on the preceding $M$ voltage values.

\begin{figure}[t]
    \centering
    \subfigure[]{
    	\begin{minipage}[b]{1\linewidth}%占比
        \centering
        \includegraphics[scale=0.27]{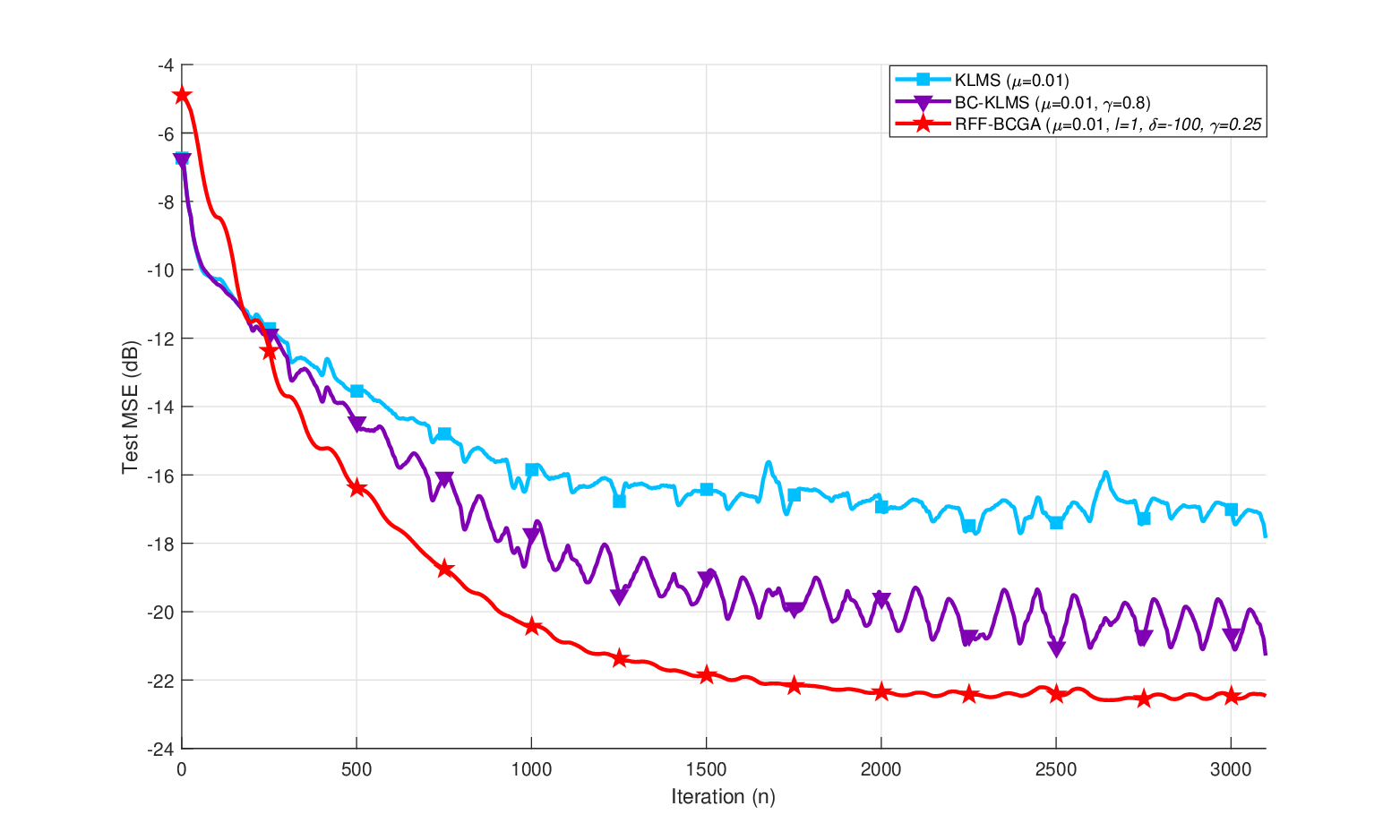} %大小 
    \end{minipage}
    }
        \subfigure[]{
    	\begin{minipage}[b]{1\linewidth}
        \centering
        \includegraphics[scale=0.27]{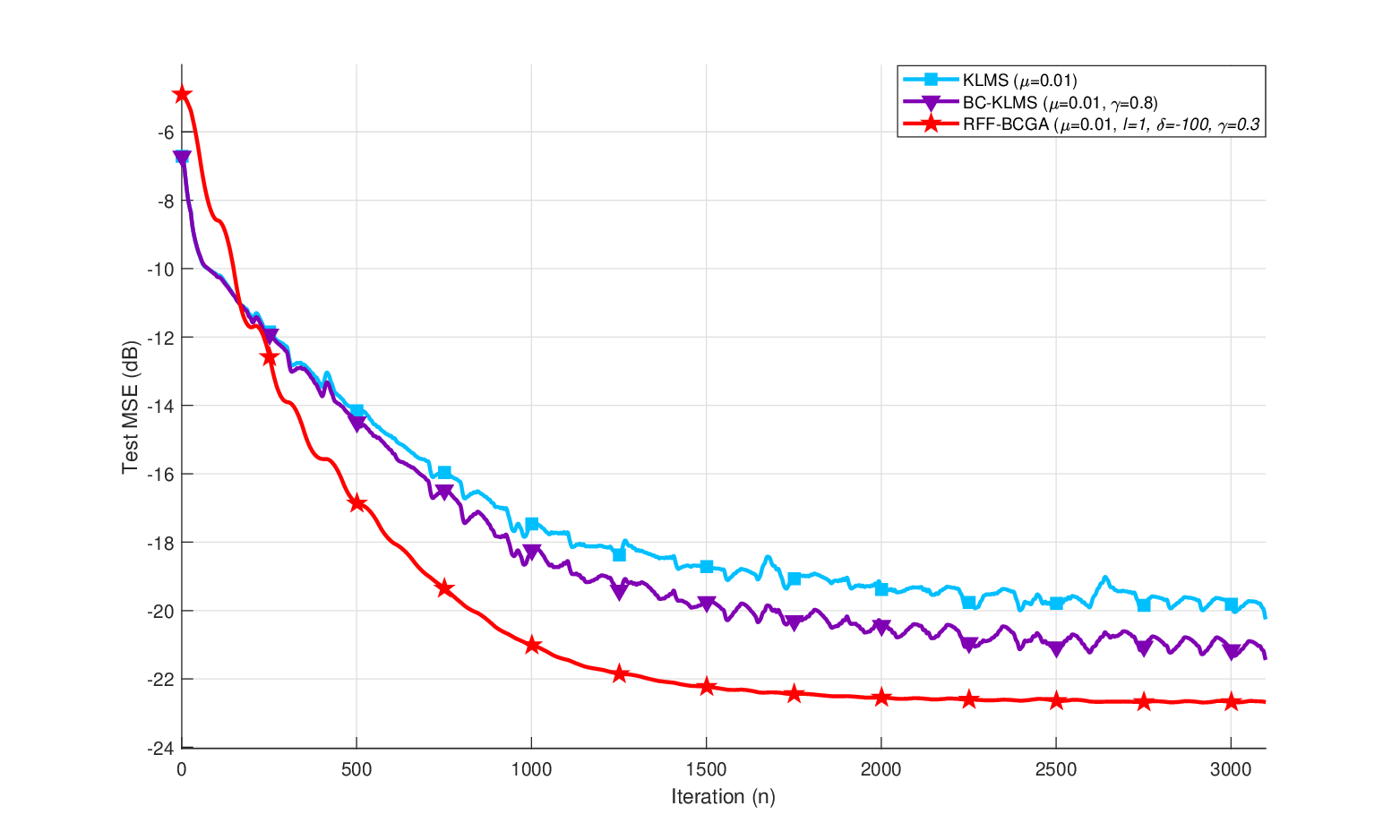}
    \end{minipage}
    }
    \caption{Testing MSE for prediction of noisy Chua's circut time series (kernel bandwidth $\xi=0.6$) under different noise conditions: (a) 5dB AWGN; (b) 10dB AWGN.}
    \label{Chua}
\end{figure} 
Fig. \ref{Chua} illustrates the prediction performance of the proposed and competing algorithms for the Chua's circuit time series when the training set is contaminated by noise. As shown in Fig. \ref{Chua}, although the same dictionary selection method is employed, the performance fluctuations of both KLMS and BCKLMS are less pronounced in this specific dataset. This observation corroborates our previous discussion, confirming that the choice of the dictionary significantly impacts algorithm performance. Furthermore, the proposed algorithm consistently achieves the best performance across different noise levels, demonstrating its robust superiority.

\subsection{Justification of Theorems 1 and 2}
\begin{figure}
\centering
\subfigure[]{\label{fig:subfig:a}
\includegraphics[width=0.48\linewidth]{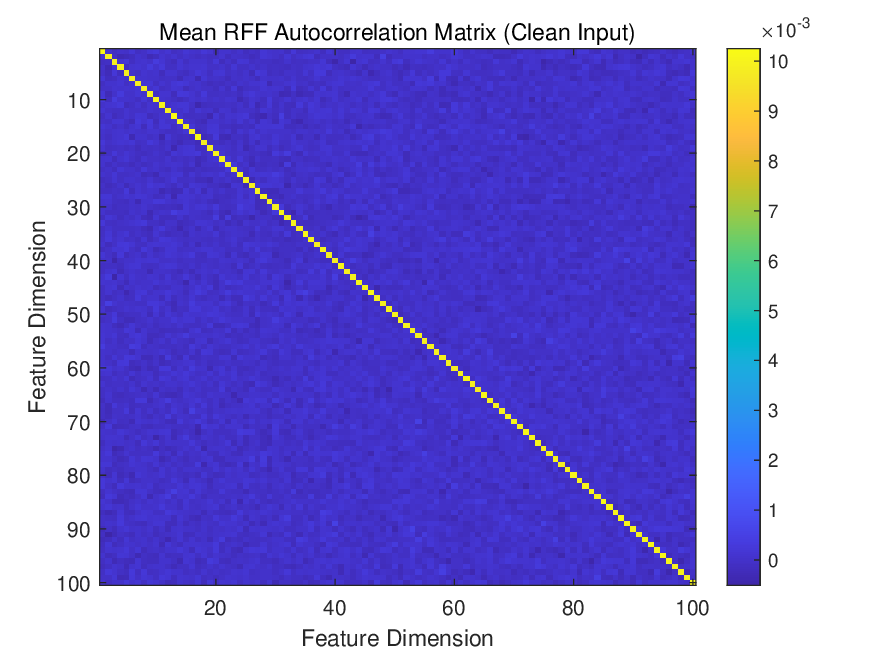}}
\hfill
\subfigure[]{\label{fig:subfig:b}
\includegraphics[width=0.48\linewidth]{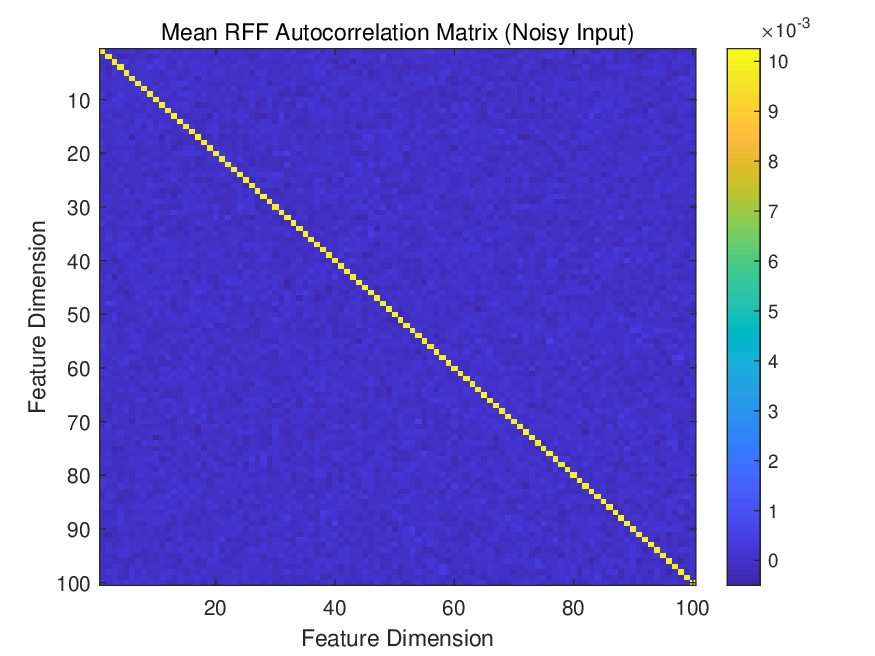}}

\subfigure[]{\label{fig:subfig:c}
\includegraphics[width=0.48\linewidth]{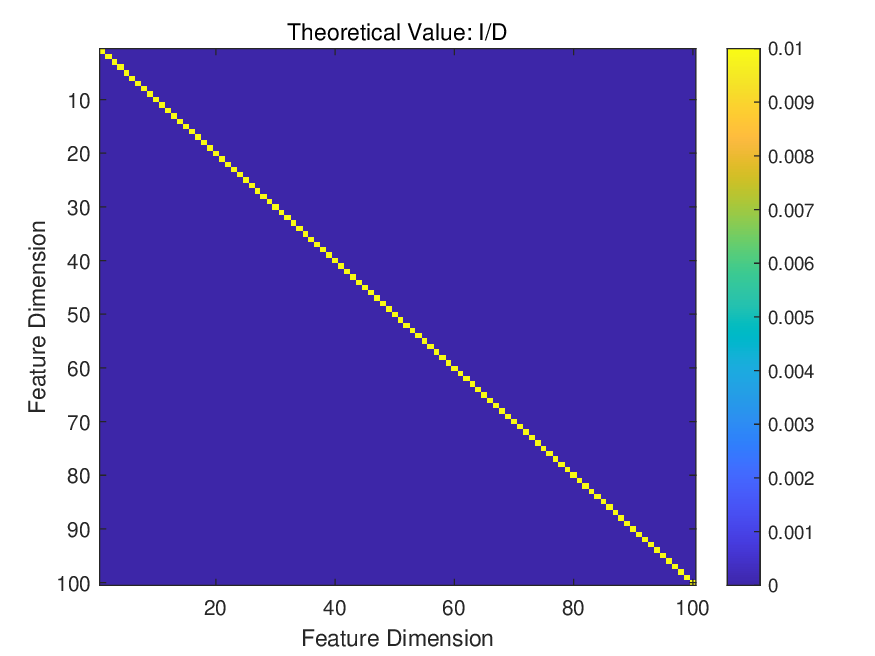}}
\hfill
\subfigure[]{\label{fig:subfig:d}
\includegraphics[width=0.48\linewidth]{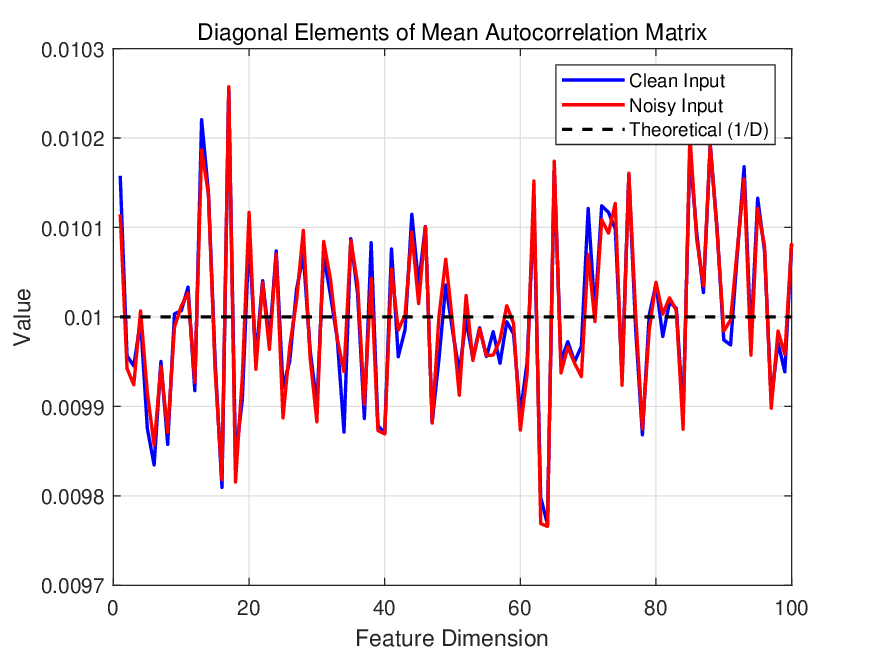}}
\caption{Simulation verification of Theorem 1 and 2}
\label{Just}
\vspace{-5mm}
\end{figure}
Fig. \ref{Just} validates the correctness of Theorems 1 and 2 over 1000 independent MC runs. We generated 100 clean Gaussian input samples with an input dimension $M=1$, the RFF dimension $ D=100$, the kernel bandwidth  $\xi=0.4$, and the AWGN $SNR$ = 10 dB. The validation was performed by comparing the simulated values of the noise-free autocorrelation matrix $\textbf{R}_i$ and the noisy autocorrelation matrix $\bar {\textbf{R}}_i$. 
As shown in Figs. \ref{Just}(a), (b), and (c), the simulated values for both matrices show excellent agreement with their theoretical counterparts. Furthermore, Fig. \ref{Just}(d) confirms the viewpoint discussed in $Remark 2$ that the noise present in the input signal has no effect on the value of the autocorrelation matrix. Consequently, $\textbf{R}_i$ and $\bar {\textbf{R}}_i$ can be treated as equivalent during the derivation process, which significantly reduces the complexity of the algorithm derivation.

\section{Conclusion}
In conclusion, this paper has addressed the limitations of existing algorithms in nonlinear EIV models by proposing the novel RFFBCGA algorithm. The proposed method maintains a fixed network dimensionality while achieving more accurate characterization of input signal features. By incorporating the flexible functional form of the GA criterion, the algorithm's robustness across diverse noise environments is significantly enhanced. We have provided theoretical analysis of both mean and mean-square performance, deriving several important theoretical findings. Extensive simulation results, including time series prediction tasks, demonstrate the superior performance of our proposed algorithm compared to existing competing methods.

\small
\bibliographystyle{IEEEtran}
\bibliography{IEEEabrv,ref}
%\newpage

\end{document}